\pdfoutput=1
\documentclass[11pt]{article}

\newif\ifreview
\reviewfalse

\ifreview
    \usepackage[review]{acl}
\else
    \usepackage[final]{acl}
\fi

\usepackage{times}
\usepackage{latexsym}

\usepackage[T1]{fontenc}
\usepackage[utf8]{inputenc}

\usepackage{microtype}

\usepackage{inconsolata}

\usepackage{graphicx}

\usepackage{custom}

\title{Towards Global AI Inclusivity: 
\\
A Large-Scale Multilingual Terminology Dataset (GIST)}

\author{\bf
Jiarui Liu,$^1$\thanks{Equal contribution.}
Iman Ouzzani,$^1$\samethanks{}
Wenkai Li,$^1$\samethanks{}
Lechen Zhang,$^2$
Tianyue Ou,$^1$\\
\bf
Houda Bouamor,$^1$
Zhijing Jin,$^{3,4,5}$
Mona Diab$^1$\\
$^1$Carnegie Mellon University, $^2$University of Michigan, $^3$University of Toronto\\
$^4$Max Planck Institute for Intelligent Systems, Tuebingen, Germany\\
$^5$Vector Institute\\
\texttt{\{jiaruil5, iouzzani, wenkail\}@andrew.cmu.edu},\\
\texttt{zjin@cs.toronto.edu}, \texttt{mdiab@andrew.cmu.edu}
}

\newcommand{\ourdataset}{\textsc{GIST}}

\begin{document}
\maketitle
\begin{abstract}

The field of machine translation has achieved significant advancements, yet domain-specific terminology translation, particularly in AI, remains challenging. We introduce \ourdataset{}, a large-scale multilingual AI terminology dataset containing 5K terms extracted from top AI conference papers spanning 2000 to 2023. The terms are translated into Arabic, Chinese, French, Japanese, and Russian using a hybrid framework that combines LLMs for extraction with human expertise for translation. The dataset's quality is benchmarked against existing resources, demonstrating superior translation accuracy through crowdsourced evaluation. \ourdataset{} is integrated into translation workflows using post-translation refinement methods that require no retraining, where LLM prompting consistently improves BLEU, COMET, and other scores. A web demonstration on the ACL Anthology platform highlights its practical application, showcasing improved accessibility for non-English speakers. This work aims to address critical gaps in AI terminology resources and fosters global inclusivity and collaboration in AI research.

\end{abstract}

\section{Introduction}

\begin{figure}[t]
    \centering
    \includegraphics[width=\linewidth]{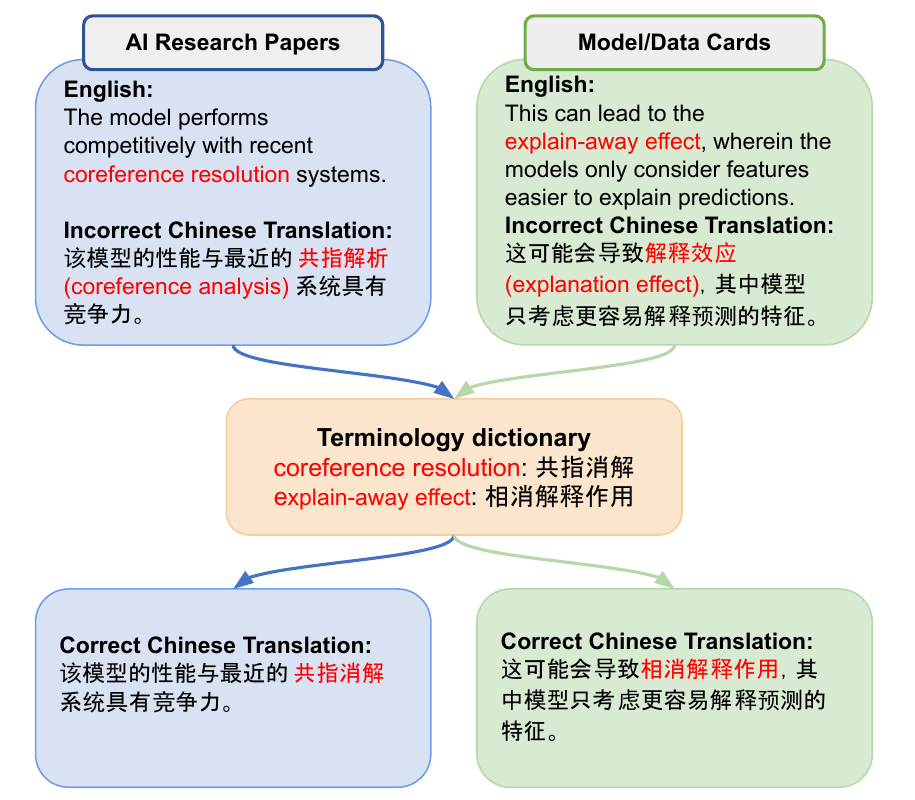}
    \caption{Direct translations of AI research papers and model cards using Google Translate generally offer fair quality but often fail to accurately translate AI-specific terminologies, potentially causing confusion or misunderstanding for readers. Our work addresses this issue by providing high-quality translations for a wide range of such terms, which can be efficiently integrated post-hoc to enhance the initial translations.}
    \label{fig:current_issues}
\end{figure}

\begin{figure*}[t]
    \centering
    \includegraphics[width=\linewidth]{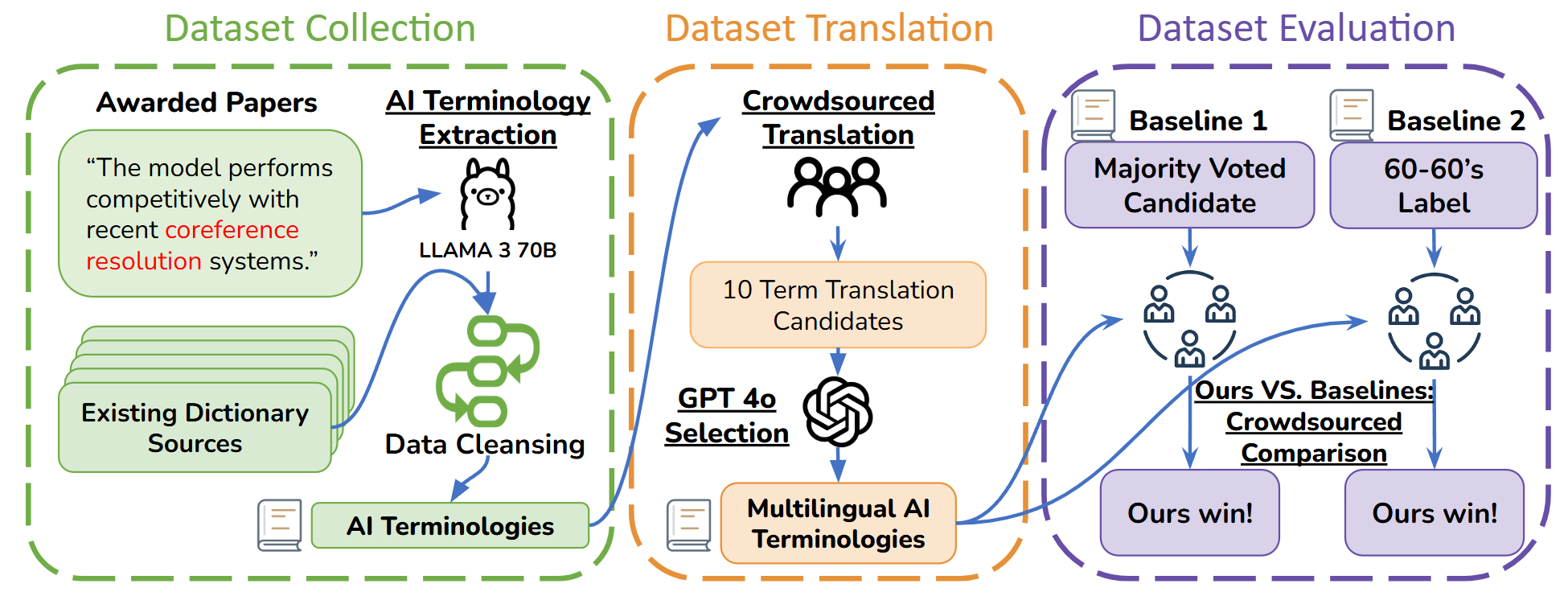}
    \caption{Overview of \ourdataset{} creation. AI terminology are extracted from awarded papers using an LLM and then combined with existing terminology dictionaries. After data cleansing, translations into five languages are generated via crowdsourcing, and an LLM is used to select the best candidate translation. Dataset quality is evaluated against two baselines: majority-vote-selected translations and translations from the 60-60 evaluation set, demonstrating the superior quality of \ourdataset{}.}
    \label{fig:overview}
\end{figure*}

The field of machine translation has made significant progress, with state-of-the-art models excelling across diverse tasks \citep{brown1990statistical, wu2016google, goyal-etal-2022-flores, haddow-etal-2022-survey} and demonstrating effectiveness in translating between high-resource and low-resource languages \citep{yao-wan-2020-multimodal, costa2022no, ranathunga2023neural}. Despite these successes, translating domain-specific scientific texts remains a persistent challenge, particularly for terminology translation \citep{cabre2010terminology, shuttleworth2014dictionary, naveen2024overview}. General-purpose translation systems often falter in accurately translating specialized terminology, leading to loss of critical details \citep{dagan1994termight, haque2020analysing}, or worse yet possibly leading to misinterpretations \citep{chmutina2021lost, yue-etal-2024-translating}.

In the field of AI, terms such as ``Coreference Resolution'' or ``Explain-Away Effect,'' are frequently translated incorrectly or inconsistently, undermining comprehension for global researchers and practitioners \citep{khuwaileh2011terminology, tehseen2018corpus}, as illustrated in \cref{fig:current_issues}. For non-English readers, who comprise a substantial portion of the global population both within and beyond the AI community \citep{ammon2003global, ding2023voices}, such inaccuracies in AI research papers and blog translation hinder access to essential knowledge, stifling research innovation and collaboration across linguistic boundaries \citep{amano2023manifold, bahji2023exclusion}. The implications extend to AI resource documentation, such as model and data cards hosted on platforms such as Hugging Face,\footnote{\href{https://huggingface.co/docs/hub/en/model-cards}{huggingface.co/docs/hub/en/model-cards}, \href{https://huggingface.co/docs/hub/en/datasets-cards}{huggingface.co/docs/hub/en/datasets-cards}} where errors in translated terminology can lead to misunderstandings or incorrect usage of models and datasets. This highlights the urgent need for precise and standardized multilingual AI terminology resources to support equitable access to AI knowledge \citep{ahuja2023mega, liu-etal-2024-automatic}.\footnote{See \cref{appn:additional_motivation} for a more in-depth discussion of our motivation.}

% \begin{figure}[t]
%     \centering
%     \includegraphics[width=\columnwidth]{fig/wordcloud.pdf}
%     \caption{WordCloud of terms tokenized with NLTK \citep{loper2002nltk} from \ourdataset{}.
%     \textcolor{red}{Z: this looks a bit less technical, more like a blog post. Do we want to keep this form?}
%     \jiarui{Haha not sure but I think your suggestions make sense! Removed this figure for now.}
%     }
%     \label{fig:wordcloud}
% \end{figure}

Existing efforts to address this gap have relied heavily on manual curation of specialized terminology by domain experts, a process that is time-consuming, resource-intensive, and difficult to scale \citep{wang2013perspectives, freitag2021experts}. Meanwhile, large language models (LLMs) have shown promise in automated terminology translation \citep{feng-etal-2024-ladder}, while their outputs often misalign with human expert standards \citep{zhang2024good}, and different models often yield inconsistent translations \citep{banik2019assembling, prieto2021translating}. The ACL 60-60 initiative in 2022 curated and translated AI-specific terms, and showcased potential for multilingual scientific communication and AI inclusivity \citep{salesky-etal-2023-evaluating}.\footnote{\href{https://2022.aclweb.org/dispecialinitiative.html}{2022.aclweb.org/dispecialinitiative.html}} However, its small scale and narrow scope %, and translation quality limitations 
underscore the need for a more comprehensive multilingual AI terminology resource at scale.

To address this, we introduce \textbf{G}lossary of Multilingual A\textbf{I} \textbf{S}cientific \textbf{T}erminology (\ourdataset{}), the first large-scale multilingual AI terminology dataset, compiling 5K AI-specific terms from award-winning papers at 18 top-tier AI conferences (2000–2023). Using a hybrid approach combining LLMs for extraction and verification with human expertise for translation, we provide high-quality translations into Arabic, Chinese, French, Japanese, and Russian \citep{Abel2013TheDO}.

To enhance accessibility of our work, we integrate the curated terminology into machine translation pipelines without requiring model retraining. We evaluate three post-translation refinement methods: prompting, word alignment and replacement, and constrained decoding. Our experiments reveal that the prompting method effectively integrates the curated dictionary, consistently improving translation quality as measured by BLEU \citep{papineni-etal-2002-bleu}, COMET \citep{rei2020comet}, ChrF \citep{popovic-2015-chrf}, ChrF++ \citep{popovic-2017-chrf}, and TER \citep{snover-etal-2006-study}. Additionally, we develop a website demonstration that incorporates our dataset into translations of ACL Anthology research papers, showcasing its practical application for non-English speakers in the AI field.

The contributions of this work are summarized as follows:
\begin{itemize}[noitemsep, left=2pt, topsep=0pt]
    \item We create \ourdataset{}, the first large-scale multilingual AI terminology dataset, with 5K terms from 879 awarded top AI conference papers (2000–2023) and broad domain coverage;\footnote{
    \ifreview
        Our data and code have been uploaded to the submission system, and will be open-sourced upon paper acceptance.
    \else
        Our code and data is available at \url{https://github.com/jiarui-liu/MultilingualAITerminology}.
    \fi
    }
    \item We develop an effective and efficient translation framework combining LLMs and human expertise to translate English terms into five languages, achieving high-quality results validated by automatic and human evaluations.
    \item To enhance accessibility, we explore three approaches to integrate the curated terminology into machine translation pipelines without retraining, and develop a website to showcase its practical use for AI paper translation.
\end{itemize}

By addressing the critical gap in multilingual AI terminology resources, this work contributes a robust solution to support equitable access to AI knowledge, fostering inclusivity and collaboration in the global research community.

\section{Related Work}

\paragraph{Multilingual Terminology Datasets}

The ACL 60-60 evaluation sets represent an important effort in translating multilingual terminology from AI presentations, but are limited to just over 250 terms \citep{salesky-etal-2023-evaluating}. 

Scientific terminology datasets are typically built using two approaches. The first  relies on human multilingual experts for extracting and translating terminology \citep{awadh2024challenges, kim-etal-2024-kpopmt}. In cases of limited expert availability, human crowdsourcing with aggregation techniques has shown excellent results \citep{zaidan2011crowdsourcing, chan2023post}. The second approach uses ML and machine translation tools for automatic collection and translation, including log-likelihood comparisons \citep{haque2018termfinder}, machine translation-based data synthesis \citep{haque2020terminology, fernando2020data, manzini2022mapping, moslem2023domain}, platform-based terminology linkers \citep{arcan-etal-2014-identification}, classifier training \citep{jin-etal-2013-mining, schumann-martinez-alonso-2018-automatic}, and prompting LLMs \citep{nishio2024extraction, shamsabadi2024large}.

However, fully human-driven approaches can be costly for large-scale multilingual datasets, while fully automated ones often lack accuracy \citep{giguere2023leveraging}. To address this, our framework integrates LLM-based extraction, human filtering, human translation, LLM validation, and merging with existing dictionaries, efficiently and effectively expanding existing terminology datasets.

\paragraph{Integrating Domain Terminology into Machine Translation}

Integrating newly collected domain terminology into machine translation systems has led to a variety of research efforts. One common approach involves training methods, such as augmenting training data with input-output pairs that include terminology for supervised fine-tuning \citep{dinu2019training, niehues-2021-continuous}, or modifying model architectures to enhance terminology awareness \citep{dinu2019training, conia2024towards}. However, these training-based approaches are inefficient for adapting to new terminology, limiting their real-world applicability.

To tackle this, terminology-aware decoding methods have emerged as a more flexible alternative. These methods, which include variants of constrained beam search \citep{anderson2016guided, hokamp2017lexically, chatterjee2017guiding, hasler2018neural, post2018fast}, slightly compromise translation accuracy for greater adaptability. Another strategy is post-hoc editing of generated translations, which typically employs word alignment techniques to identify and replace term translations in the output \citep{zenkel2019adding, chen2020accurate, ferrando2022towards}. Recently, LLMs have been leveraged to integrate expected terminology directly into the translations \citep{bogoychev-chen-2023-terminology}. Our work explores multiple terminology integration approaches, including constrained beam search, decoding logits adjustment, word alignment and replacement, and refinement through prompting.

\section{\ourdataset{} Dataset Construction}

% basic dataset information: dataset size, languages included, what each example includes, section details.

We construct \ourdataset{}, a dataset comprising around 5K English AI terminology and their translations into Arabic, Chinese, French, Japanese, and Russian. The basic lexical statistics of the dataset is presented in \cref{tab:data_stats}. Additional dataset statistics are presented in \cref{appn:additional_data_stats}. %We detail our terminology collection process in \cref{sec:term_collection} and the translation procedure in \cref{sec:term_translation}.

\begin{table}[h]
\centering
    \setlength\tabcolsep{1.5pt}
\resizebox{\columnwidth}{!}{
\begin{tabular}{lccccc}
\toprule
 & Arabic & Chinese & French & Japanese & Russian \\
\midrule
\# Terms & 4,844 & 6,426 & 6,527 & 4,770 & 5,167 \\
Unique En Words & 2,470 & 3,244 & 3,470 & 2,424 & 2,615 \\
Unique Tgt Words & 3,161 & 2,838 & 4,036 & 2,050 & 4,210 \\
En Words/Term & 2.02 & 2.05 & 2.07 & 2.02 & 2.01 \\
Tgt Words/Term & 2.36 & 2.26 & 2.68 & 2.53 & 2.16 \\
En Chars/Term & 16.99 & 17.26 & 17.44 & 16.96 & 16.94 \\
Tgt Chars/Term & 15.22 & 4.66 & 21.27 & 6.89 & 20.20 \\
\bottomrule
\end{tabular}
}
\caption{Statistics of the dataset across languages. ``En'' denotes English, and ``Tgt'' denotes the target language. Statistics with standard deviations are presented in \cref{tab:data_stats_full}.}
\label{tab:data_stats}
\end{table}

\subsection{Terminology Curation}
\label{sec:term_collection}

Our dataset follows the ACL 60-60 initiative that aims to collect scientific terminology in the AI field. We source AI terminology from two primary channels: AI research papers published online and existing AI terminology dictionaries. For the research papers, our objective is to compile a substantial number of terms from high-quality AI papers spanning a long time frame. To identify representative papers, a natural approach would be to crawl the most cited papers using online search engines or platforms. However, no tools or APIs currently enable this. Instead, we focus on awarded papers announced on the websites of top AI conferences. Specifically, we collect all awarded papers, spanning awards such as Best Paper, Outstanding Paper, and other recognitions from the venues listed in \cref{tab:top_ai_conference}, covering the years 2000 to 2023. This approach ensures comprehensive coverage of recognized research work. In total, we collect 879 paper PDFs from arXiv and other online repositories and process them into text files using SciPDF.\footnote{\url{https://github.com/titipata/scipdf_parser}} As we analyze later in \cref{sec:dataset_coverage}, this collection strategy provides broad terminology and domain coverage of AI terminology.

As supported by previous research \citep{xu2024large, dagdelen2024structured, shamsabadi2024large}, recent LLMs such as \texttt{LLaMA 3} \citep{dubey2024llama} have demonstrated strong capabilities in scientific terminology extraction tasks. While the definition of ``AI'' lacks a clear and universally agreed boundary, LLMs are trained on vast datasets that reflect human knowledge, enabling them to classify terms as AI-related based on their contextual relevance. Accordingly, we leverage \texttt{LLaMA-3-70B-Instruct} to extract AI terminology from award-winning papers, providing specific instructions to guide the extraction process.

To define AI-specific terminology, we impose the following criteria: (1) the term must be a noun or noun phrase, (2) it should be specialized to AI, encompassing core concepts, methods, models, algorithms, or systems, and (3) it should have either no meaning or a distinct meaning outside the AI domain. We process the text in sentence chunks of up to 64 words to stay within \texttt{LLaMA}'s optimal context-handling capabilities. Additionally, for each unique term, we record up to three different contexts during extraction to ensure sufficient contextual diversity.

After extraction, we perform multiple quality assurance steps. We remove terms that appear in only one paper to ensure representativeness. Moreover, we exclude abbreviations and terms starting with special characters, and filter out non-noun phrases. Duplicates are eliminated, and \texttt{GPT-4o} \citep{hurst2024gpt} is employed to further refine the list by filtering out non-AI terms based on the same criteria. Finally, three AI domain experts review the terms to remove any remaining unqualified entries. To enhance the dataset, we also integrate terminology from external sources, including the 60-60 initiative dataset, government websites, Wikipedia, and other online resources. Consequently, the number of English terms varies across languages in \ourdataset{}. Full details of the terminology collection process are provided in \cref{appn:term_collection_details}.

\subsection{Terminology Translation}
\label{sec:term_translation}

% discussion of the difference among the chosen languages
In selecting target languages for translation, we aim to encompass a range of morphological complexities and varying levels of resource availability, particularly in AI-related publications. First, Chinese and Japanese exhibit minimal morphological variation in nouns and noun phrases, relying primarily on word order and context rather than inflection. In contrast, Arabic, French, and Russian are morphologically complex, characterized by extensive inflectional systems. Second, French, Chinese, Japanese, and Russian were selected due to the presence of large native-speaking scientific communities \citep{Ammon2012, chahal2022country}, which may have historically contributed to the development of well-established AI terminology. In contrast, Arabic, despite being widely spoken, may lack the same depth of scientific vocabulary, particularly in rapidly evolving fields such as AI. This disparity underscores the potential for our approach to contribute to the standardization and development of cohesive terminology across different languages in AI publications.

\begin{table*}[ht]
\centering
\resizebox{\textwidth}{!}{
\begin{tabular}{lcccc}
\toprule
 & A. Both translations are good & B. Method 1 translation is better & C. Method 2 translation is better & D. Both translations are bad \\
 \midrule
Arabic & 739, 45.76\% & 461, 28.54\% & 329, 20.37\% & 72, 4.46\% \\
Chinese & 468, 50.59\% & 266, 28.76\% & 163, 17.62\% & 25, 2.70\% \\
French & 438, 48.67\% & 274, 30.44\% & 170, 18.89\% & 17, 1.89\% \\
Japanese & 587, 56.99\% & 251, 24.37\% & 159, 15.44\% & 27, 2.62\% \\
Russian & 626, 54.43\% & 348, 30.26\% & 150, 13.04\% & 14, 1.22\% \\
 \bottomrule
\end{tabular}
}
\caption{Distribution of annotators' choices for AI terminology ratings in Task 1, comparing the GPT-4o-selected candidate (Method 1) with the majority-voted candidate (Method 2). The table reports the total counts and corresponding ratios for each choice.}
\label{tab:mturk_rating_distribution_gpt4o}
\end{table*}

To evaluate whether state-of-the-art LLMs can effectively perform terminology translation as a generation task, we initially experiment with two advanced LLMs and one API: \texttt{Claude 3 Sonnet} \citep{anthropic2024claude3}, \texttt{GPT-3.5-Turbo} \citep{openai2023devday}, and Google Translate API \citep{wu2016google}. We measure the agreement among the three methods using exact match. However, as shown in \cref{tab:mt_agreement_ratio}, the three-model agreement ratio was only around 15\% for most languages except Chinese, and the two-model agreement ratio was only about 40\%. These results reveal significant inconsistencies in the AI terminology translations produced by these systems, and highlight the need for human input to ensure reliable AI terminology translations. The detailed procedure and the prompt used for these translations is presented in \cref{appn:term_translation_details}.

Given these findings, we opt for human annotation to ensure translation accuracy. To achieve this, we utilize Amazon Mechanical Turk (MTurk) for crowdsourced annotations. A demonstration of the MTurk task is shown in \cref{fig:mturk_generation}. We instruct participants to take on the task only if they specialized in AI and are fluent in both English and one of the target languages. Annotators are tasked with generating accurate translations for each AI terminology, with relevant contexts provided and the terminology highlighted in yellow in context. To maintain quality, we implement a rigorous qualification process. Annotators are first tested on a toy set of 10 carefully selected AI terms, and only those who perform well are allowed to proceed with the full task. Additionally, we monitor submissions daily and filtered out participants who provide random or low-quality translations during the annotation process. All in all, for each term, we collect 10 annotations per target language.

Finally, we use \texttt{GPT-4o} to select the best translation from the annotators’ submissions and Google API Translation for each term, ensuring high-quality results for our final dataset. As analyzed in \cref{sec:dataset_assess_q1}, leveraging \texttt{GPT-4o} is crucial for maintaining the quality of the translations.

\section{Dataset Quality Assessments}
\label{sec:data_quality}

To thoroughly evaluate the translation quality in \ourdataset{}, we conduct two additional crowdsourced rating tasks: (1) In \cref{sec:dataset_assess_q1}, we investigate whether using \texttt{GPT-4o} is necessary to select the best translation candidate; (2) In \cref{sec:dataset_assess_q2}, we compare the quality of our translation with the evaluation set from the 60-60 initiative.

\subsection{Task 1: Is an LLM Necessary for Selecting the Best Translation Candidate?}
\label{sec:dataset_assess_q1}

To ensure the selection of the best candidate from the annotators' generations, we explore two methods for candidate selection. The first method only uses \texttt{GPT-4o} to select the best translation candidate among 10 annotations and one Google Translation for all terms. The second method relies on majority voting among the 11 translations: a translation candidate is selected if it appears in more than 5 out of 11 annotations. In cases where no majority is reached, \texttt{GPT-4o} is prompted to select the best translation candidate. 

To evaluate these approaches, we conduct a separate MTurk task, involving a different group of participants, to compare the two methods on a randomly sampled subset of approximately 200 terms per language. Participants are asked to choose one of four options: (A) Both translations are good; (B) Method 1 translation is better; (C) Method 2 translation is better; (D) Both translations are bad. For each language, we collect 5 annotations per term. As shown in \cref{tab:mturk_rating_distribution_gpt4o}, \texttt{GPT-4o}'s candidate selection consistently outperforms majority voting across all five languages. These results highlight the necessity of \texttt{GPT-4o} in achieving high-quality translations for crowdsourced annotations.

However, human involvement remains indispensable. As discussed in \cref{sec:term_translation}, state-of-the-art LLMs and machine translation systems fail to provide consistent answers, rendering automatic translation through agreement across multiple models infeasible. Thus, by combining the expertise of LLMs as verifiers with humans as input sources, we achieve efficient, accurate, and reliable translations.

\begin{table*}[ht]
\centering
\resizebox{\textwidth}{!}{
\begin{tabular}{lcccc}
\toprule
 & A. Both translations are good & B. Method 1 translation is better & C. Method 2 translation is better & D. Both translations are bad \\
 \midrule
Arabic & 376, 46.42\% & 238, 29.38\% & 143, 17.65\% & 46, 5.68\% \\
Chinese & 197, 37.17\% & 228, 43.02\% & 85, 16.04\% & 17, 3.21\% \\
French & 152, 39.48\% & 168, 43.64\% & 53, 13.77\% & 10, 2.60\% \\
Japanese & 295, 57.28\% & 162, 31.46\% & 36, 6.99\% & 21, 4.08\% \\
Russian & 172, 39.09\% & 198, 45.00\% & 38, 8.64\% & 25, 5.68\% \\
 \bottomrule
\end{tabular}
}
\caption{Distribution of annotators' choices for AI terminology ratings in Task 2, comparing the translations in our dataset (Method 1) with those in the 60-60 initiative evaluation set (Method 2). The table reports the total counts and corresponding ratios for each choice.}
\label{tab:mturk_rating_distribution_6060}
\end{table*}

\subsection{Task 2: Is Our Dataset Translation Better than 60-60?}
\label{sec:dataset_assess_q2}

The 60-60 initiative dataset also focuses on AI terminology translations and includes the five languages we consider. This overlap allows us to intersect the English terms in our dataset with those in the 60-60 evaluation dataset and compare their respective translations. We retrieve the translations from both datasets and conduct a comparative assessment through a crowdsourced evaluation, same as the process detailed in \cref{sec:dataset_assess_q1}.

The results presented in \cref{tab:mturk_rating_distribution_6060} show that the translations in our dataset consistently and significantly outperform those from the 60-60 dataset across all five languages. Additionally, as shown in \cref{tab:mturk_rating_agreement}, annotators demonstrated fair agreement when rating the Arabic translations, and moderate agreement, with Fleiss' Kappa values ranging from 0.4 to 0.5, for translations in the other four languages. These findings underscore the superior quality of translations in our dataset compared to the 60-60 dataset. Refer to \cref{appn:dataset_quality_assessment_details} for further details on the assessment of the two tasks.

\section{Dataset Coverage Assessment}
\label{sec:dataset_coverage}

\paragraph{Domain Coverage}

We first investigate the AI domains covered by the collected terms in \ourdataset{} and their distributions. We use \texttt{GPT-4o-mini} \citep{hurst2024gpt} to identify the specific AI domains for each term, following the taxonomy proposed by \citet{ding2023voices}, which clusters the research domains of AI scholars. \Cref{fig:domain_coverage} shows the distribution of the top six AI domains in \ourdataset{}, with terms most frequently categorized into statistics, mathematics, computer science (CS), natural language processing (NLP), data science (DS), and computer vision (CV). We further embed all AI terminology in \ourdataset{} using the \texttt{all-MiniLM-L6-v2} model \citep{reimers-2019-sentence-bert} and apply Uniform Manifold Approximation and Projection (UMAP) \citep{mcinnes2018umap} for dimensionality reduction to a two-dimensional space. The visualization in \cref{fig:domain_umap} reveals that terms in domains such as NLP, CV, statistics, and mathematics form distinct clusters, while terms in CS and DS are more dispersed.

\paragraph{Terminology Coverage}

We surmise that our awarded paper dataset provides essential coverage of the terms in our terminology dictionary, where many terms are actually repeatedly extracted multiple times from different awarded papers. We present the rarefaction curve in \cref{fig:rarefaction_curve}, which visually demonstrates the relationship between the number of extracted terms and the dataset size. 

To statistically validate our claim, we conduct a one-sided one-sample t-test to compare the mean coverage ratio of the terminology dictionary. Specifically, we test whether the mean coverage ratio of randomly sampled subsets, constituting 60\% of the terminology dictionary, is significantly above 80\%. The null hypothesis (\(H_0\)) assumes that the mean coverage ratio of these subsets is less than or equal to 80\%, while the alternative hypothesis (\(H_a\)) assumes that the mean coverage ratio exceeds 80\%. Using 1,000 random samples, the analysis yields a t-statistic of 64.78 and a p-value of 0, rejecting the null hypothesis. These results provide statistical evidence supporting the sufficiency of our dataset in covering the terminology dictionary. Refer to \cref{appn:dataset_coverage_assessment_details} for more details.

\setlength{\abovecaptionskip}{6pt}
\begin{table*}[h]
\centering
\resizebox{\linewidth}{!}{

\begin{tabular}{ll|c@{\hspace{2pt}}c@{\hspace{2pt}}c|c@{\hspace{2pt}}c@{\hspace{2pt}}c|c@{\hspace{2pt}}c@{\hspace{2pt}}c|c@{\hspace{2pt}}c@{\hspace{2pt}}c|c@{\hspace{2pt}}c@{\hspace{2pt}}c}
\toprule
\multirow{2}{*}{\textbf{Model}} & \multirow{2}{*}{\textbf{Metric}} & \multicolumn{3}{c|}{\textbf{Arabic}} & \multicolumn{3}{c|}{\textbf{Chinese}} & \multicolumn{3}{c|}{\textbf{French}} & \multicolumn{3}{c|}{\textbf{Japanese}} & \multicolumn{3}{c}{\textbf{Russian}} \\

& & \textbf{D} & \textbf{+P} & \textbf{+W} & \textbf{D} & \textbf{+P} & \textbf{+W} & \textbf{D} & \textbf{+P} & \textbf{+W} & \textbf{D} & \textbf{+P} & \textbf{+W} & \textbf{D} & \textbf{+P} & \textbf{+W} \\

\midrule

\multicolumn{17}{l}{\textbf{Evaluation Set: 60-60}} \\
\specialrule{1.2pt}{0pt}{2.5pt}

\multirow{2}{*}{\texttt{aya-expanse}} & BLEU & 20.11 &  \textcolor{ForestGreen}{+ 1.23} & \textcolor{ForestGreen}{+ 0.18}  & 27.31 & \textcolor{ForestGreen}{+ 1.33} & \textcolor{ForestGreen}{+ 0.24}  & 33.05 & \textcolor{ForestGreen}{+ 2.46} & \textcolor{ForestGreen}{+ 0.20}  & 14.59 & \textcolor{ForestGreen}{+ 0.61} & \textcolor{ForestGreen}{+ 0.32}  & 16.59 & \textcolor{ForestGreen}{+ 1.59} & \textcolor{red}{- 0.05}  \\

& COMET & 81.96 & \textcolor{ForestGreen}{+ 0.71} & \textcolor{red}{- 0.52}  & 83.43 & \textcolor{ForestGreen}{+ 1.57} & \textcolor{ForestGreen}{+ 0.08}  & 81.83 & \textcolor{ForestGreen}{+ 1.06} & \textcolor{red}{- 0.11}  & 88.54 & \textcolor{ForestGreen}{+ 0.32} & \textcolor{red}{- 0.01}  & 82.27 & \textcolor{ForestGreen}{+ 0.69} & \textcolor{red}{- 2.02}  \\

\cline{2-17}

\multirow{2}{*}{\texttt{aya-23-8B}} & BLEU & 19.98 & \textcolor{ForestGreen}{+ 0.54} & \textcolor{red}{- 0.21}  & 26.08 & \textcolor{ForestGreen}{+ 0.47} & \textcolor{ForestGreen}{+ 0.39}  & 33.85 & \textcolor{ForestGreen}{+ 2.28} & \textcolor{red}{- 0.11}  & 15.06 & \textcolor{ForestGreen}{+ 0.87} & \textcolor{ForestGreen}{+ 0.36}  & 15.77 & \textcolor{ForestGreen}{+ 1.05} & \textcolor{ForestGreen}{+ 0.37}  \\
& COMET & 84.02 & \textcolor{ForestGreen}{+ 0.81} & \textcolor{red}{- 0.24}  & 85.12 & \textcolor{ForestGreen}{+ 0.58} & \textcolor{ForestGreen}{+ 0.38}  & 82.40 & \textcolor{ForestGreen}{+ 0.94} & \textcolor{red}{- 0.15}  & 87.92 & \textcolor{ForestGreen}{+ 0.50} & \textcolor{ForestGreen}{+ 0.09}  & 81.91 & \textcolor{ForestGreen}{+ 0.40} & \textcolor{red}{- 2.26}  \\

\cline{2-17}

\multirow{2}{*}{\texttt{gpt-4o-mini}} & BLEU & 23.58 & \textcolor{ForestGreen}{+ 1.07} & \textcolor{red}{- 0.00}  & 32.64 & \textcolor{ForestGreen}{+ 1.60} & \textcolor{ForestGreen}{+ 0.66}  & 40.80 & \textcolor{ForestGreen}{+ 3.08} & \textcolor{ForestGreen}{+ 0.50}  & 21.46 & \textcolor{ForestGreen}{+ 0.64} & \textcolor{ForestGreen}{+ 0.19}  & 17.25 & \textcolor{ForestGreen}{+ 1.07} & \textcolor{red}{- 0.13}  \\
& COMET & 85.77 & \textcolor{ForestGreen}{+ 0.69} & \textcolor{red}{- 0.44}  & 87.30 & \textcolor{ForestGreen}{+ 0.48} & \textcolor{ForestGreen}{+ 0.26}  & 84.56 & \textcolor{ForestGreen}{+ 0.68} & \textcolor{red}{- 0.04}  & 89.96 & \textcolor{ForestGreen}{+ 0.14} & \textcolor{ForestGreen}{+ 0.01}  & 83.68 & \textcolor{ForestGreen}{+ 0.38} & \textcolor{red}{- 2.29}  \\

\cline{2-17}

\multirow{2}{*}{\texttt{nllb}} & BLEU & 22.38 & \textcolor{ForestGreen}{+ 1.37} & \textcolor{ForestGreen}{+ 0.64}  & 17.29 & \textcolor{ForestGreen}{+ 1.92} & \textcolor{ForestGreen}{+ 1.02}  & 34.93 & \textcolor{ForestGreen}{+ 2.86} & \textcolor{ForestGreen}{+ 0.21}  & 6.19 & \textcolor{ForestGreen}{+ 2.42} & \textcolor{ForestGreen}{+ 0.53}  & 17.30 & \textcolor{ForestGreen}{+ 1.54} & \textcolor{ForestGreen}{+ 1.07}  \\
& COMET & 83.52 & \textcolor{ForestGreen}{+ 0.83} & \textcolor{red}{- 0.45}  & 78.22 & \textcolor{ForestGreen}{+ 2.95} & \textcolor{ForestGreen}{+ 0.73}  & 82.83 & \textcolor{ForestGreen}{+ 1.00} & \textcolor{red}{- 0.19}  & 77.82 & \textcolor{ForestGreen}{+ 3.80} & \textcolor{ForestGreen}{+ 0.39}  & 81.41 & \textcolor{ForestGreen}{+ 0.97} & \textcolor{red}{- 1.54}  \\

\cline{2-17}

\multirow{2}{*}{\texttt{seamless}} & BLEU & 23.13 & \textcolor{ForestGreen}{+ 1.16} & \textcolor{red}{- 0.03}  & 26.26 & \textcolor{ForestGreen}{+ 0.97} & \textcolor{ForestGreen}{+ 0.80}  & 40.04 & \textcolor{ForestGreen}{+ 2.08} & \textcolor{red}{- 0.57}  & 14.56 & \textcolor{ForestGreen}{+ 0.74} & \textcolor{ForestGreen}{+ 0.05}  & 17.18 & \textcolor{ForestGreen}{+ 1.71} & \textcolor{ForestGreen}{+ 1.17}  \\
& COMET & 84.07 & \textcolor{ForestGreen}{+ 0.94} & \textcolor{red}{- 0.38}  & 83.44 & \textcolor{ForestGreen}{+ 1.48} & \textcolor{ForestGreen}{+ 0.50}  & 83.86 & \textcolor{ForestGreen}{+ 0.78} & \textcolor{red}{- 0.07}  & 85.05 & \textcolor{ForestGreen}{+ 1.06} & \textcolor{ForestGreen}{+ 0.16}  & 82.33 & \textcolor{ForestGreen}{+ 0.56} & \textcolor{red}{- 1.87}  \\

\midrule

\multicolumn{17}{l}{\textbf{Evaluation Set: AI Papers \& Model Cards}} \\
\specialrule{1.2pt}{0pt}{2.5pt}

\multirow{2}{*}{\texttt{aya-expanse}} & BLEU & 11.47 & \textcolor{ForestGreen}{+ 0.37} & \textcolor{ForestGreen}{+ 0.10}  & 12.04 & \textcolor{ForestGreen}{+ 0.94} & \textcolor{ForestGreen}{+ 0.17}  & 18.84 & \textcolor{ForestGreen}{+ 1.71} & \textcolor{red}{- 0.85}  & 8.11 & \textcolor{red}{- 0.03} & \textcolor{ForestGreen}{+ 0.04}  & 13.84 & \textcolor{ForestGreen}{+ 0.21} & \textcolor{ForestGreen}{+ 0.32}  \\
& COMET & 80.98 & \textcolor{ForestGreen}{+ 0.46} & \textcolor{red}{- 0.51}  & 82.42 & \textcolor{ForestGreen}{+ 0.56} & \textcolor{red}{- 0.01}  & 81.16 & \textcolor{ForestGreen}{+ 0.36} & \textcolor{red}{- 0.79}  & 85.48 & \textcolor{ForestGreen}{+ 0.34} & \textcolor{ForestGreen}{+ 0.07}  & 82.76 & \textcolor{ForestGreen}{+ 0.47} & \textcolor{red}{- 2.10}  \\

\cline{2-17}

\multirow{2}{*}{\texttt{aya-23-8B}} & BLEU & 14.28 & \textcolor{ForestGreen}{+ 0.81} & \textcolor{ForestGreen}{+ 0.28}  & 14.50 & \textcolor{ForestGreen}{+ 0.39} & \textcolor{ForestGreen}{+ 0.19}  & 24.49 & \textcolor{ForestGreen}{+ 2.36} & \textcolor{ForestGreen}{+ 0.08}  & 9.22 & \textcolor{ForestGreen}{+ 0.23} & \textcolor{ForestGreen}{+ 0.35}  & 16.39 & \textcolor{ForestGreen}{+ 1.36} & \textcolor{ForestGreen}{+ 0.61}  \\
& COMET & 81.55 & \textcolor{ForestGreen}{+ 1.03} & \textcolor{red}{- 0.62}  & 83.88 & \textcolor{ForestGreen}{+ 0.68} & \textcolor{ForestGreen}{+ 0.04}  & 82.55 & \textcolor{ForestGreen}{+ 1.22} & \textcolor{red}{- 0.70}  & 84.42 & \textcolor{ForestGreen}{+ 0.81} & \textcolor{ForestGreen}{+ 0.02}  & 82.72 & \textcolor{ForestGreen}{+ 1.14} & \textcolor{red}{- 2.08}  \\

\cline{2-17}

\multirow{2}{*}{\texttt{gpt-4o-mini}} & BLEU & 14.37 & \textcolor{ForestGreen}{+ 0.53} & \textcolor{red}{- 0.42}  & 17.21 & \textcolor{ForestGreen}{+ 1.22} & \textcolor{ForestGreen}{+ 1.39}  & 24.45 & \textcolor{ForestGreen}{+ 4.38} & \textcolor{ForestGreen}{+ 1.28}  & 10.55 & \textcolor{ForestGreen}{+ 0.05} & \textcolor{ForestGreen}{+ 0.03}  & 18.02 & \textcolor{ForestGreen}{+ 1.52} & \textcolor{ForestGreen}{+ 0.40}  \\
& COMET & 83.56 & \textcolor{ForestGreen}{+ 0.86} & \textcolor{red}{- 0.19}  & 86.08 & \textcolor{ForestGreen}{+ 0.25} & \textcolor{red}{- 0.13}  & 84.75 & \textcolor{ForestGreen}{+ 0.33} & \textcolor{red}{- 1.03}  & 87.91 & \textcolor{ForestGreen}{+ 0.16} & \textcolor{red}{- 0.01}  & 84.92 & \textcolor{ForestGreen}{+ 0.44} & \textcolor{red}{- 2.02}  \\

\cline{2-17}

\multirow{2}{*}{\texttt{nllb}} & BLEU & 15.42 & \textcolor{red}{- 0.31} & \textcolor{red}{- 0.77}  & 10.24 & \textcolor{ForestGreen}{+ 2.19} & \textcolor{ForestGreen}{+ 2.07}  & 22.68 & \textcolor{ForestGreen}{+ 2.73} & \textcolor{ForestGreen}{+ 0.90}  & 8.24 & \textcolor{ForestGreen}{+ 1.08} & \textcolor{ForestGreen}{+ 0.88}  & 19.18 & \textcolor{red}{- 0.10} & \textcolor{red}{- 0.24}  \\
& COMET & 81.23 & \textcolor{ForestGreen}{+ 1.28} & \textcolor{red}{- 0.50}  & 80.19 & \textcolor{ForestGreen}{+ 1.61} & \textcolor{ForestGreen}{+ 0.38}  & 78.70 & \textcolor{ForestGreen}{+ 3.81} & \textcolor{red}{- 1.59}  & 83.05 & \textcolor{ForestGreen}{+ 1.71} & \textcolor{ForestGreen}{+ 0.70}  & 80.46 & \textcolor{ForestGreen}{+ 2.80} & \textcolor{red}{- 2.38}  \\

\cline{2-17}

\multirow{2}{*}{\texttt{seamless}} & BLEU & 15.38 & \textcolor{ForestGreen}{+ 1.09} & \textcolor{ForestGreen}{+ 0.45}  & 13.67 & \textcolor{ForestGreen}{+ 1.10} & \textcolor{ForestGreen}{+ 0.73}  & 24.34 & \textcolor{ForestGreen}{+ 5.21} & \textcolor{ForestGreen}{+ 1.49}  & 9.42 & \textcolor{ForestGreen}{+ 0.56} & \textcolor{ForestGreen}{+ 0.42}  & 18.43 & \textcolor{ForestGreen}{+ 0.95} & \textcolor{ForestGreen}{+ 0.35}  \\
& COMET & 81.96 & \textcolor{ForestGreen}{+ 1.18} & \textcolor{red}{- 0.39}  & 80.70 & \textcolor{ForestGreen}{+ 2.18} & \textcolor{ForestGreen}{+ 0.16}  & 83.76 & \textcolor{ForestGreen}{+ 0.97} & \textcolor{red}{- 0.94}  & 83.70 & \textcolor{ForestGreen}{+ 0.88} & \textcolor{ForestGreen}{+ 0.10}  & 83.12 & \textcolor{ForestGreen}{+ 1.50} & \textcolor{red}{- 1.79}  \\

\bottomrule
\end{tabular}

}
\caption{Evaluation results across five models and five languages using BLEU and COMET metrics. The first black value in each column represents the direct translation score (\textbf{D}). The second and third values, shown in \textcolor{red}{red} and \textcolor{ForestGreen}{green}, indicate the relative performance change when applying the prompting-powered refinement method (\textbf{P}) and the word alignment method (\textbf{W}), respectively, compared to direct translation. See \cref{tab:exp_results_full} for the complete results of additional metrics and ablations.}

\label{tab:exp_results}
\end{table*}

\section{Experiments}

We explore terminology integration approaches that do not need model retraining. We quantitatively evaluate two methods: terminology refinement via post-hoc LLM prompting and terminology substitution guided by word alignment.

\subsection{Methods}

\paragraph{Terminology Refinement: LLM Prompting}

With the recent advancements in multilingual and instruction-following capabilities of LLMs, it is now possible to leverage these models to refine translations. Specifically, we prompt the LLMs to revise the initial translation produced by a machine translation model, incorporating relevant term translations from our terminology dictionary within the provided context. For this refinement, we employ \texttt{GPT-4o-mini} and use the prompt illustrated in \cref{fig:term_refinement_prompt}. Refer to \cref{appn:term_refinement_prompting_details} for more details.

\paragraph{Terminology Substitution: Word Alignment}

We apply the word alignment approach introduced by \citet{dou2021word} to identify and substitute term translations within the output. We use the multilingual BERT base model \citep{DBLP:journals/corr/abs-1810-04805} to tokenize both source and target sentences, and then process these inputs through its hidden layers to produce contextualized embeddings. We determine alignments by computing dot-product similarities between source and target token embeddings. High-confidence alignments are then filtered using a threshold of 1e-4, and subword alignments are aggregated to generate word-level mappings. Lastly, we directly replace identified term translations with those from \ourdataset{}. We also conduct a post-hoc prompting after word alignment using \texttt{GPT-4o-mini} to render the translations morphologically coherent and correct. Refer to \cref{appn:term_word_alignment_details} for more details.  Additionally, we conduct a qualitative evaluation of traditional decoding techniques, including constrained beam search and token-level logits adjustment. See \cref{appn:term_decoding} for more details.

\subsection{Experiment Setup}

\paragraph{Evaluation Set} We conduct experiments on two evaluation sets: the \textbf{60-60 Evaluation Set} and the \textbf{AI Papers and Model Cards Evaluation Set}. The first evaluation set is from the 60-60 initiative, providing terminology translations into five target languages. As detailed in \cref{sec:dataset_assess_q2}, our analysis shows that our dataset achieves higher quality compared to the 60-60 set. Motivated by this, we enhance the labels of the 60-60 evaluation set by replacing AI terminology in their translations with \texttt{GPT-4o} while maintaining grammatical correctness and ensuring no loss in translation quality. This process generates updated ground truth labels for the evaluation.

The second evaluation set is manually created by combining text from two sources: 50 held-out AI research papers and 50 model cards generated by \citet{liu-etal-2024-automatic}. From this set, we randomly sampled 500 English text chunks from it for evaluation. To create ground truth labels for this set, we use Google Translate, a state-of-the-art machine translation model \citep{zhu2023multilingual, santosa2024integration}, to produce initial translations. We then prompt \texttt{GPT-4o} to refine these translations by updating AI terminology. This evaluation set is designed to explore the application of our AI terminology dictionary in two major domains: AI research papers and model cards.

\begin{figure*}[h]
    \centering
    \includegraphics[width=0.95\linewidth]{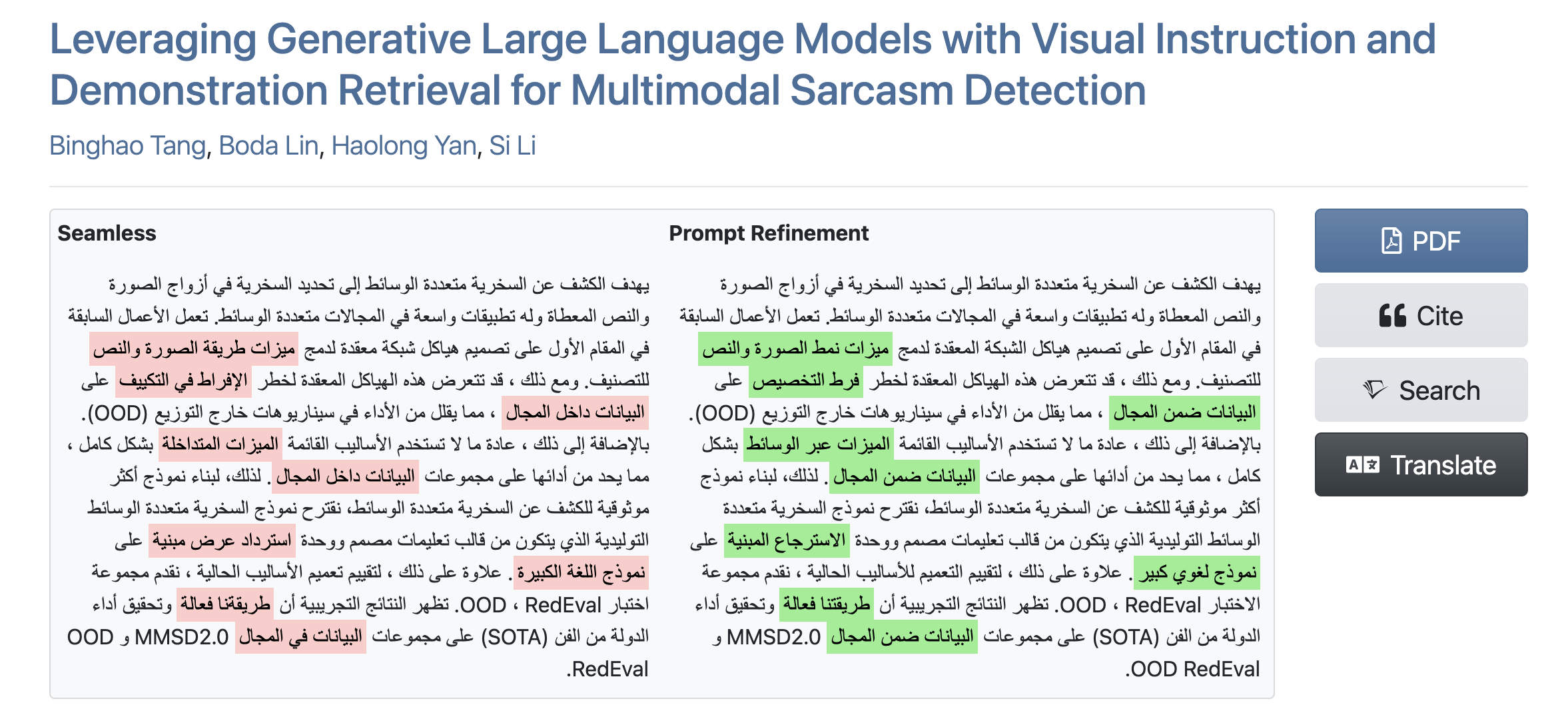}
    \caption{The interface demonstrates a paper example from \citet{tang-etal-2024-leveraging} and introduces a new translation feature on the ACL Anthology website. Non-English speakers can click the ``Translate'' button at the bottom right of the webpage and select their preferred language for translation. The page dynamically displays two translations side by side: one using direct translation from a machine translation model (e.g., \texttt{seamless} in the figure) and the other enhanced with prompting-powered refinement applied to the left translation. Highlighted text indicates updated terminology integrated from \ourdataset{}, showcasing improved translation performance.}
    \label{fig:web_demo}
\end{figure*}

% Both evaluation set references have been verified by human experts and exhibit high quality. 
To validate the quality of our evaluation set references, we conducted a manual evaluation involving five expert annotators who assessed 100 randomly sampled examples (50 from each evaluation set). Each annotator was presented with the source English sentence, its translation, and the relevant AI terminologies and their translations. They completed three tasks: (1) rating overall translation quality on a scale from 1 (nonsensical) to 5 (excellent), (2) assessing grammatical and morphological correctness (Yes/No), and (3) evaluating the accuracy of terminology translation (Yes/No). As summarized in \cref{tab:manual_eval}, results indicate high annotation consistency and translation quality across all five languages. Average translation scores ranged from 3.11 (Russian) to 3.96 (Chinese), and binary accuracy for grammar and terminology typically exceeded 80\%. These findings demonstrate the robustness and linguistic reliability of our evaluation sets and affirm the quality of \texttt{GPT-4o}-assisted refinement.
Further details are provided in \cref{appn:evaluation_set_creation_details}.

\vspace{-0.4em}
\paragraph{Models} We evaluate the following models: \texttt{gpt-4o-mini} \citep{hurst2024gpt}, \texttt{hf-seamless-m4t-large} \citep{barrault2023seamless}, \texttt{nllb-200-3.3B} \citep{costa2022no}, \texttt{aya-23-8B} \citep{aryabumi2024aya}, and \texttt{aya-expanse-8B} \citep{dang2024aya}. While the prompting method applies to all models, the word alignment method is not applicable to \texttt{gpt-4o-mini}, as it does not provide access to its model weights.

\paragraph{Evaluation Metrics} We adopt BLEU \citep{papineni-etal-2002-bleu}, COMET \citep{rei2020comet}, ChrF \citep{popovic-2015-chrf}, ChrF++ \citep{popovic-2017-chrf}, and TER metrics \citep{snover-etal-2006-study} for the quantitative evaluation of translation quality, following the methodology of \citet{salesky-etal-2023-evaluating}. We utilize the \texttt{wmt22-comet-da} model \citep{rei-etal-2022-comet} to compute COMET scores.

\subsection{Experiment Results}
\label{subsec:exp_results}

% qualitative results: manual evaluation, the generation speed is 100x slower than prompting refinement and term substitution.

\cref{tab:exp_results} presents the quantitative evaluation results under the experimental settings described earlier. We draw several key observations:

First, the results across the two evaluation sets show consistency, indicating the robustness of the findings. Among the models, \texttt{gpt-4o-mini} achieves the best overall translation performance. Among the remaining four models, differences in performance are less pronounced.

Several trends emerge when comparing different terminology integration approaches. The prompting-powered refinement method consistently outperforms direct translation across nearly all languages, models, and evaluation metrics, which highlights its effectiveness in incorporating AI-specific terminology into translations. 

The word alignment method demonstrates mixed performance, showing improvements for Chinese and Japanese translations but leading to declines for Arabic, French, and Russian. This discrepancy is due to linguistic differences: Chinese exhibits minimal morphological changes, allowing straightforward substitution of terminology with limited disruption to surrounding syntax; In contrast, languages like Arabic often require agreement in gender, number, and syntactic roles, making noun replacement more complex and error-prone. These findings underscore the superior performance of the prompt-powered refinement method, as well as the importance of tailoring terminology integration approaches to the linguistic characteristics of target languages. See \cref{appn:additional_quantitative_res,appn:additional_qualitative_res} for further discussion of the results.

We further validate our findings using a one-sided paired t-test in \cref{appn:stat_test_res}, which confirms that the prompt-powered refinement method significantly outperforms both direct translation and the word alignment approach across all evaluated languages.

\vspace{-0.4em}
\section{Website Demonstration}

To facilitate real-world usage of our AI terminology dictionary, we built on the 60-60 initiative by modifying the ACL Anthology website layout and introducing a new terminology translation feature,\footnote{\url{https://acl6060.org/}} as illustrated in \cref{fig:web_demo}. The ACL Anthology website was chosen for this demonstration due to its extensive collection of AI-related research papers, making it an ideal platform to showcase the potential impact of our work. In this demonstration, translations of key terms are refined and standardized based on the terminology dictionary we developed, providing more accurate terminology translations.%\footnote{
% \ifreview
%     Our website demo will be available upon acceptance.
% \else
%     Our code for the website is available at \url{https://github.com/ImanOu123/acl-anthology-mod}.
% \fi
% }
This enhancement represents a step toward improving access to AI knowledge for non-English speakers by offering a consistent and reliable translation system, and broadens the accessibility of AI research to a global audience.

\vspace{-0.3em}
\section{Conclusion}

We present \ourdataset{}, a large-scale multilingual AI terminology dataset addressing gaps in translating AI-specific terms. Combining LLM-based extraction and validation with human expertise, it includes 5K English terms with translations into five languages, surpassing the ACL 60-60 benchmark. LLM prompting proved effective for post-hoc terminology integration, improving translation quality across five metrics. We also provide a website demonstration to enhance the accessibility of our work for non-English speakers, supporting equitable AI knowledge access and fostering global collaboration in AI research.

\section*{Limitations}

This work is subject to several limitations. First, our dataset assumes a one-to-one correspondence between English terms and their translations, which does not account for cases where multiple equally valid translations exist for a single term. This simplification may overlook the nuanced variations in AI terminology usage across languages.

Despite significant efforts in collecting terminologies, our coverage is not exhaustive. The field of artificial intelligence lacks a well-defined boundary, making it challenging to ensure comprehensive inclusion of all relevant AI terms. Additionally, while we provide translations for five widely used languages, this represents only a subset of the global linguistic diversity and leaves many other languages unaddressed.

% The evaluation set was created using Google Translate and \texttt{GPT-4o}, tools whose translation accuracy has not been verified in our work by human experts. This reliance might introduce potential errors in the evaluation data, affecting its reliability.

Furthermore, while our methodology is tailored for AI terminology translation, its application to other domains may require adaptation. Although our LLM + Human hybrid framework for data collection, translation, and evaluation is broadly applicable and does not rely on AI-specific models or tools, domain-specific terminology translation poses unique challenges, such as variations in terminology collection and the need for domain-specific expertise in evaluation.

In addition, expanding the terminology dataset automatically presents unique challenges. The rapidly evolving nature of the AI field requires frequent updates to ensure the GIST dataset remains current and comprehensive. At present, identifying new, domain-relevant terms still depends heavily on human expertise to assess whether a term is both novel and impactful within the AI research community. While large language models can aid in extracting terminology and generating candidate terms, their utility is constrained by fixed knowledge cutoffs, limiting their ability to capture the most recent developments. As a result, fully automating the update process remains a complex and unresolved task.

Nonetheless, we hope this work inspires the community to further advance the creation and refinement of multilingual AI terminology dictionaries, addressing these limitations and extending coverage to more languages and domains.

\section*{Ethical Considerations}

We manually collected awarded papers from official conference websites. Only papers available under open-source licenses for research use were downloaded. Similarly, model card contents used in this study were sourced from openly shared materials by their respective authors.

During data processing, we ensured that no personal information, such as human names irrelevant to AI methods or metrics, was included in the dataset. Automatic and manual reviews were conducted to verify the exclusion of any sensitive or private details.

For term translation, we employed LLMs to assist with extraction and validation. While we acknowledge that LLMs may exhibit biases when selecting the best translations, manual evaluations confirmed the superior performance of LLMs in this task.

In crowdsourced experiments, we respected participant privacy by not collecting any demographic information. All contributors were fairly compensated according to MTurk's payment standards. Additionally, our collected terminology dataset does not involve ethically sensitive or controversial content, focusing exclusively on technical terms relevant to AI.

% read other machine translation paper dicsussion on this
% especially for terminology collection and translation

\ifreview
\else
    \section*{Acknowledgments}
    This material is based in part upon work supported by the German Federal Ministry of Education and Research (BMBF): Tübingen AI Center, FKZ: 01IS18039B; by the Machine Learning Cluster of Excellence, EXC number 2064/1 – Project number 390727645; by Schmidt Sciences SAFE-AI Grant; by NSERC Discovery Grant RGPIN-2025-06491; by a National Science Foundation award (\#2306372); by a Swiss National Science Foundation award (\#201009) and a Responsible AI grant by the Haslerstiftung.
    
    We extend our gratitude to the Minister of State for Artificial Intelligence of the United Arab Emirates for making the Arabic AI terminology dictionary publicly available. We thank DataFranca for granting us access to their French AI terminology dictionary. We thank Jiqizhixin for sharing their Chinese AI terminology database. We also thank the Russian Association of Artificial Intelligence for providing access to the Russian AI terminology dictionary. For more details, please refer to \cref{appn:term_collection_details}.
\fi
%Datafranca

\bibliography{custom}

\appendix

\section{Further Discussion on AI Terminology Translation}
\label{appn:additional_motivation}

\begin{figure*}[t]
    \centering
    \includegraphics[width=\linewidth]{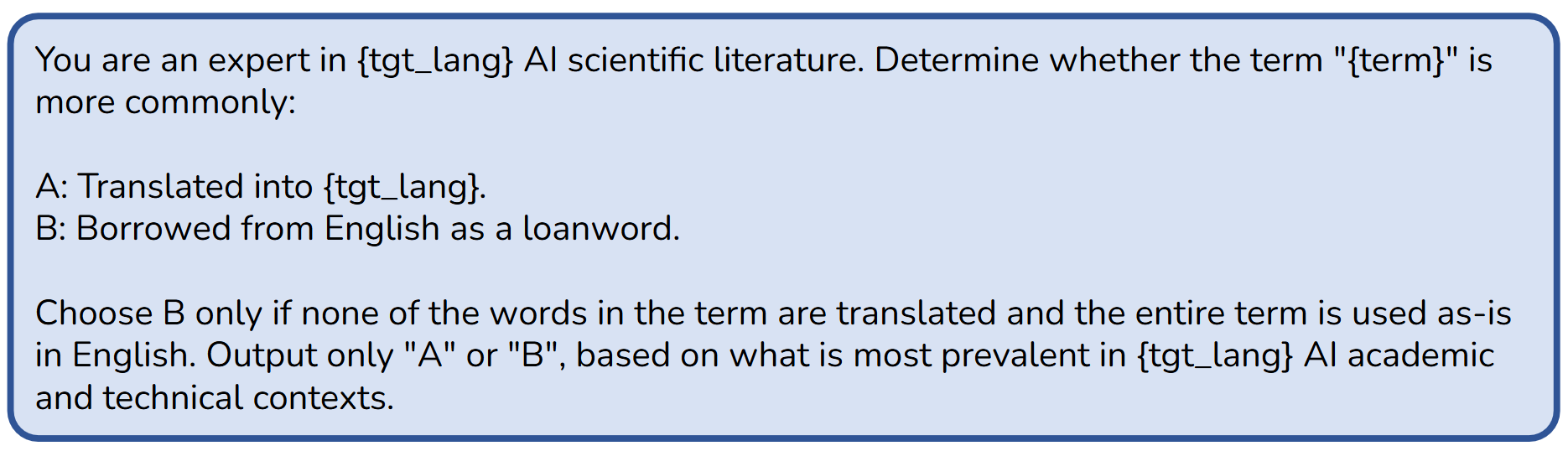}
    \caption{Prompt used to determine whether GIST terms should be translated.}
    \label{fig:prompt_motivation_loanwords}
\end{figure*}

In many non-English texts nowadays, AI terminology is often borrowed as untranslated loanwords. However, translating these terms into local languages is crucial for the following reasons:

\begin{enumerate}
\item Translation democratizes AI for non-English speakers. Translating AI terminology into native languages plays a vital role in making the field more accessible to non-English-speaking communities. For those outside the AI field, translating these terms fosters a clearer understanding, as it is easier for them to grasp complex concepts in their own language rather than through borrowed English terms. Prior studies highlight that translating English loanwords into Arabic and other languages helps to fill lexical gaps \citep{calabrese2009loan, al2024arabicization}, making terms more meaningful while preserving linguistic and cultural identities. This practice promotes the standardization of AI terminology, reduces dependency on foreign languages, and improves educational accessibility for non-English speakers.
\item Our work aligns with and extends the ACL 60-60 initiative \citep{salesky-etal-2023-evaluating}, which aims to translate AI-specific terms to enhance multilingual scientific communication and foster AI inclusivity. By expanding this effort, we seek to improve the scale, scope, and quality of AI terminology translation, furthering the global reach of AI research.
\item The prevalence of untranslated loanwords varies significantly across languages. The prevalence of untranslated loanwords is more common in languages influenced by English or with linguistic similarities to English. However, this is not the case for many other languages. As a concession to this variability, we employed \texttt{GPT-4o-mini} to determine whether terms in the GIST dataset should be translated. \cref{tab:motivation_loanwords} presents the results, based on the prompt in \cref{fig:prompt_motivation_loanwords}, which highlights the disparities across languages.
\end{enumerate}

\begin{table}[h]
    \centering
    \begin{tabular}{lc}
        \toprule
        Language & \shortstack{Percentage of Terms \\ Requiring Translation} \\
        \midrule
        Chinese  & 95.0\% \\
        Arabic   & 73.0\% \\
        French   & 72.0\% \\
        Japanese & 53.8\% \\
        Russian  & 80.5\% \\
        \bottomrule
    \end{tabular}
    \caption{Proportion of AI terms requiring translation across different languages.}
    \label{tab:motivation_loanwords}
\end{table}

\section{Additional Dataset Details}
\label{sec:appendix}

\subsection{Additional Dataset Statistics}
\label{appn:additional_data_stats}

\begin{table*}[t]
\centering
\resizebox{0.8\linewidth}{!}{
\begin{tabular}{lccccc}
\toprule
 & Arabic & Chinese & French & Japanese & Russian \\
\midrule
\# Terms & 4844 & 6426 & 6527 & 4770 & 5167 \\
Unique En Words & 2470 & 3244 & 3470 & 2424 & 2615 \\
Unique Tgt Words & 3161 & 2838 & 4036 & 2050 & 4210 \\
En Words/Term & $2.02 \pm 0.59$ & $2.05 \pm 0.68$ & $2.07 \pm 0.67$ & $2.02 \pm 0.58$ & $2.01 \pm 0.59$ \\
Tgt Words/Term & $2.36 \pm 0.83$ & $2.26 \pm 0.90$ & $2.68 \pm 1.19$ & $2.53 \pm 0.98$ & $2.16 \pm 0.80$ \\
En Chars/Term & $16.99 \pm 5.97$ & $17.26 \pm 6.60$ & $17.44 \pm 6.57$ & $16.96 \pm 5.91$ & $16.94 \pm 5.90$ \\
Tgt Chars/Term & $15.22 \pm 5.66$ & $4.66 \pm 1.96$ & $21.27 \pm 8.49$ & $6.89 \pm 3.16$ & $20.20 \pm 7.83$ \\
\bottomrule
\end{tabular}
}
\caption{Lexical statistics of the dataset across languages, including standard deviations. ``En'' denotes English, and ``Tgt'' denotes the target language. Terms are tokenized into words using \texttt{nltk.word\_tokenize} for English, Arabic, French, and Russian, \texttt{jieba} for Chinese, and the \texttt{MeCab Owakati} tokenizer for Japanese.}
\label{tab:data_stats_full}
\end{table*}

To better understand the composition and characteristics of the dataset, we performed a comprehensive statistical analysis across multiple dimensions, including domain distribution, semantic clustering, and lexical structure across five languages: Arabic, Chinese, French, Japanese, and Russian. Below, we present key findings through detailed visualizations and lexical statistics.

\paragraph{Lexical Statistics.}
Table~\ref{tab:data_stats_full} summarizes the lexical characteristics of the dataset across the five languages. Key metrics include:
\begin{itemize}
    \item \textbf{Number of Terms:} The dataset contains 4,844 to 6,527 terms per language, with French and Chinese having the largest repositories.
    \item \textbf{Unique Words:} English terms comprise 2,400--3,400 unique words across datasets. Target languages exhibit varying lexical diversity, with French and Russian showing higher uniqueness due to linguistic richness.
    \item \textbf{Words per Term:} On average, English terms consist of approximately two words (\( \sim 2.02 \)), while target languages show higher variability. French terms, for instance, require more words (\(2.68 \pm 1.19\)), reflecting language-specific expansion during translation.
    \item \textbf{Characters per Term:} English terms maintain consistent lengths (\( \sim 17 \) characters), while target languages vary significantly. For example, Chinese terms are concise (\(4.66 \pm 1.96\)) due to its logographic script, whereas French (\(21.27 \pm 8.49\)) and Russian (\(20.20 \pm 7.83\)) terms are longer, reflecting the morphology of these languages.
\end{itemize}

To tokenize terms, we utilized \texttt{nltk.word\_tokenize} for English, Arabic, French, and Russian; \texttt{jieba} for Chinese; and \texttt{MeCab Owakati} for Japanese. These tools ensured language-specific tokenization accuracy, enabling detailed lexical analysis.

\paragraph{Terminology Examples} To identify and show the most frequently used terms in the original set of awarded papers, we extracted and ranked the top 150 terminologies based on their occurrence frequency. \cref{tab:terms_top150} provides a comprehensive list of these terms, categorized by their rank and grouped for clarity.

\begin{table*}[t]
\centering
\resizebox{\textwidth}{!}{
\begin{tabular}{rlrlrl}
\toprule
Rank & Terminology & Rank1 & Terminology (Continued)1 & Rank2 & Terminology (Continued)2 \\
\midrule
1 & Algorithm & 51 & Latent space & 101 & Recall \\
2 & Model & 52 & Node & 102 & Q-learning \\
3 & Classifier & 53 & Transfer learning & 103 & Lasso \\
4 & Transformer & 54 & Stochastic gradient descent & 104 & Transition matrix \\
5 & Machine learning & 55 & Feature vector & 105 & Linear regression \\
6 & Policy & 56 & Gibbs sampling & 106 & Meta-learning \\
7 & Learning rate & 57 & Baseline & 107 & Segmentation \\
8 & Neural network & 58 & Generative model & 108 & Fourier transform \\
9 & Language model & 59 & Ontology & 109 & Epoch \\
10 & Loss function & 60 & Attention & 110 & Learning algorithm \\
11 & Reinforcement learning & 61 & Training set & 111 & Topic model \\
12 & Encoder & 62 & Data mining & 112 & Time complexity \\
13 & Deep learning & 63 & Manifold & 113 & Feature selection \\
14 & Decoder & 64 & Discriminator & 114 & Knowledge distillation \\
15 & Gradient descent & 65 & F1 score & 115 & Word embedding \\
16 & Beam search & 66 & Dynamic programming & 116 & Euclidean distance \\
17 & Machine translation & 67 & Adam optimizer & 117 & Covariance \\
18 & Computer vision & 68 & Eigenvalue & 118 & Hyper-parameter \\
19 & Dataset & 69 & Vector & 119 & Test set \\
20 & Graph & 70 & State-of-the-art & 120 & Attention mechanism \\
21 & Markov chain & 71 & Regularization & 121 & Oracle \\
22 & Kernel & 72 & Backpropagation & 122 & Question answering \\
23 & Marginal likelihood & 73 & Greedy algorithm & 123 & Point cloud \\
24 & Objective function & 74 & Optical flow & 124 & Local minima \\
25 & Gradient & 75 & Mutual information & 125 & N-gram \\
26 & Reward function & 76 & Weight decay & 126 & Semi-supervised learning \\
27 & Entropy & 77 & Posterior distribution & 127 & Batch normalization \\
28 & Tensor & 78 & Bounding box & 128 & Homomorphism \\
29 & Active learning & 79 & Disentanglement & 129 & Markov \\
30 & Natural language processing & 80 & Convolution & 130 & Mini-batch \\
31 & Perplexity & 81 & Semantic segmentation & 131 & Subgraph \\
32 & Posterior & 82 & Logit & 132 & Bias \\
33 & Logistic regression & 83 & Loss & 133 & Arg min \\
34 & Covariance matrix & 84 & Multi-task learning & 134 & State space \\
35 & Self-attention & 85 & Matrix & 135 & Dimensionality \\
36 & Data augmentation & 86 & Binary classification & 136 & Random variable \\
37 & Object detection & 87 & In-context learning & 137 & Gaussian distribution \\
38 & Inference & 88 & Validation set & 138 & Optimizer \\
39 & Cosine similarity & 89 & Cost function & 139 & Weight vector \\
40 & Sample complexity & 90 & Corpus & 140 & Named entity recognition \\
41 & Value function & 91 & Estimator & 141 & Kernel matrix \\
42 & Probability distribution & 92 & Lemma & 142 & Discount factor \\
43 & Generator & 93 & Parser & 143 & Hidden layer \\
44 & Adam & 94 & Feature space & 144 & Domain adaptation \\
45 & Supervised learning & 95 & Sentiment analysis & 145 & Frobenius norm \\
46 & Dropout & 96 & Token & 146 & Positional encoding \\
47 & Classification & 97 & Unsupervised learning & 147 & Seq2seq \\
48 & Ground truth & 98 & State & 148 & Cross validation \\
49 & Arg max & 99 & Infoset & 149 & Gaussian process \\
50 & K-means & 100 & Precision & 150 & Coreference resolution \\
\bottomrule
\end{tabular}
}
\caption{Top 150 terms with the highest frequency in the original set of awarded papers.}
\label{tab:terms_top150}
\end{table*}

\paragraph{Temporal Statistics} We analyze the distribution of terms in \ourdataset{} based on the publication years of the papers. \cref{tab:temporal_stats} presents the top 10 most frequent terms in awarded papers for each year. This analysis offers insights into the temporal evolution of AI terminology.

\begin{table*}[t]
\centering
\begin{tabularx}{\textwidth}{l X}
\toprule
Year & Terms \\
\midrule
2000 & feature vector, machine translation, probabilistic model, inference, loss function, feature space, BLEU, translation model, model selection, OOV \\

2001 & reinforcement learning, policy, Ablation study, baseline, machine learning, bias, learning rate, state, gradient descent, tensor \\

2002 & machine learning, machine translation, posterior distribution, inference, weight vector, training set, time complexity, tf-idf, mutual information, dot product \\

2003 & softmax, computer vision, machine learning, local minima, gradient descent, neural network, Dataset, generative model, Gaussian noise, time complexity \\

2004 & deep learning, ground truth, neural network, bounding box, 3D object detection, mean square error, machine learning, cost volume, gradient, receptive field \\

2005 & neural network, loss function, machine translation, learning rate, gradient, perplexity, Kernel, validation set, beam search, feature vector \\

2006 & NLP, machine learning, beam search, language model, Question Answering, Dataset, regularization, overfitting, greedy algorithm, neural network \\

2007 & learning rate, probability distribution, machine learning, deep learning, loss, binary classification, covariance matrix, Adam optimizer, corpus, estimator \\

2008 & machine learning, softmax, dropout, language model, computer vision, bias, baseline, neural network, F1 score, NLP \\

2009 & computer vision, language model, softmax, NLP, deep learning, learning rate, clustering, natural language processing, neural network, loss function \\

2010 & machine learning, learning rate, neural network, gradient descent, Euclidean distance, regularization, convex optimization, learning algorithm, Gaussian noise, gradient \\

2011 & learning rate, NLP, deep learning, overfitting, supervised learning, Lemma, inference, eigenvalue, baseline, Cross Entropy Loss \\

2012 & softmax, NLP, neural network, attention, deep learning, dropout, State-of-the-art, word embedding, learning rate, attention mechanism \\

2013 & machine learning, loss function, Dataset, inference, NLP, learning rate, language model, stochastic gradient descent, probability distribution, decoder \\

2014 & language model, NLP, machine learning, computer vision, convolutional layer, learning rate, deep learning, loss function, softmax, sigmoid \\

2015 & NLP, language model, learning rate, feature vector, baseline, loss function, recall, validation set, probabilistic model, Markov chain \\

2016 & learning rate, decoder, classification, loss function, NLP, Dataset, data augmentation, machine learning, State-of-the-art, beam search \\

2017 & unsupervised learning, supervised learning, neural network, reinforcement learning, Dataset, dynamic programming, inference, validation set, NLP, epoch \\

2018 & NLP, neural network, learning rate, machine learning, language model, Dataset, inference, natural language processing, loss function, BERT \\

2019 & machine learning, learning rate, NLP, neural network, softmax, gradient descent, natural language processing, computer vision, node, Dataset \\

2020 & NLP, Dataset, loss function, language model, natural language processing, learning rate, reinforcement learning, BERT, Adam optimizer, baseline \\

2021 & learning rate, NLP, natural language processing, machine learning, language model, loss function, computer vision, BERT, overfitting, decoder \\

2022 & machine learning, neural network, NLP, learning rate, computer vision, Dataset, loss function, deep learning, supervised learning, node \\

2023 & machine learning, NLP, learning rate, computer vision, deep learning, Dataset, softmax, machine translation, natural language processing, neural network \\
\bottomrule
\end{tabularx}
\caption{Top 10 most frequent terms in awarded papers for each year from 2000 to 2023.}
\label{tab:temporal_stats}
\end{table*}

\paragraph{Conclusion.}
The statistical analysis highlights the diversity and interdisciplinary nature of the dataset. Figures~\ref{fig:domain_coverage} and~\ref{fig:domain_umap} illustrate domain-wise distributions and semantic clusters, while Table~\ref{tab:data_stats_full} quantifies lexical variations across languages. Together, these findings provide a robust understanding of the dataset's structure, supporting its utility for multilingual and domain-specific AI applications.

\subsection{Terminology Collection Details}
\label{appn:term_collection_details}

We selected terms from awarded papers as this approach provides an efficient and manageable way to curate a representative sample of influential AI research. While we acknowledge that valuable AI terminology also exists in non-awarded yet influential papers, capturing all relevant terms across the vast AI field is infeasible. Nonetheless, we believe our dataset offers a comprehensive representation in terms of domain coverage and unique terminology. We considered expanding our selection to include highly cited papers; however, to our knowledge, no automated method reliably identifies such papers. Even if one existed, determining an appropriate citation threshold for inclusion would remain a challenge.

To refine our terminology selection, we employ a two-step procedure to filter out non-nominal phrases. First, we prompt \texttt{GPT-4} to retain only nouns or noun phrases as candidates. Subsequently, we use a POS tagger to further remove any phrases that do not contain a noun.

We integrate terms from the WikiPedia Glossary of AI\footnote{\href{https://en.wikipedia.org/wiki/Glossary\_of\_artificial\_intelligence}{en.wikipedia.org/wiki/Glossary\_of\_artificial\_intelligence}}, which serves as a comprehensive starting point for artificial intelligence (AI)-related terminology, ensuring alignment with globally recognized concepts and definitions. To enhance multilingual coverage and domain relevance, we also include terms from several other specialized AI terminology initiatives across different languages:

\textbf{Arabic AI Dictionary\footnote{\href{https://ai.gov.ae/ar/ai-dictionary/}{ai.gov.ae/ar/ai-dictionary/}}:} Published by the Arabic Government AI Office, this resource aims to elevate the status of the Arabic language in AI, standardize terminology, reduce linguistic ambiguity, and foster better integration of Arabic speakers into the global AI community. By providing accurate translations and clear definitions for English AI terms, this dictionary promotes knowledge dissemination and encourages collaboration within the Arabic-speaking AI ecosystem.

\textbf{Chinese GitBook AI Term Database\footnote{\href{https://jiqizhixin.gitbook.io/artificial-intelligence-terminology-database}{https://jiqizhixin.gitbook.io/artificial-intelligence-terminology-database}}:} Developed by Jiqizhixin (Machine Heart), this database represents an extensive effort to document technical terms encountered during the translation of AI articles and research papers. Starting with practical usage, the project has evolved to incorporate domain-specific expansions based on authoritative textbooks and expert input, offering the Chinese AI community a unified and precise reference for both academic and industrial applications.

\textbf{French AI Dictionary\footnote{\href{https://datafranca.org/wiki/Cat\%C3\%A9gorie:GRAND\_LEXIQUE\_FRAN\%C3\%87AIS}{https://datafranca.org/wiki/Cat\%C3\%A9gorie:\\GRAND\_LEXIQUE\_FRAN\%C3\%87AIS}}: } The first comprehensive French reference tool for data science and AI, this dictionary addresses the needs of public service, commerce, research, and education. It aims to bridge the gap between French and English AI terminologies, ensuring accessibility and standardization for French-speaking professionals and researchers.

\textbf{Russian AI Dictionary\footnote{\href{https://www.raai.org/pages/UGFnZVR5cGU6MTAwMw==}{https://www.raai.org/pages/\\UGFnZVR5cGU6MTAwMw==}}:} This initiative captures the interdisciplinary nature of AI by including terminology drawn from fields such as logic, psychology, linguistics, and cybernetics. Leveraging contributions from Russian and Soviet experts, this dictionary emphasizes the frequency and relevance of terms within AI-specific contexts, providing a culturally adapted yet globally aligned resource for Russian-speaking researchers.

\begin{figure*}[h]
    \centering
    \includegraphics[width=\linewidth]{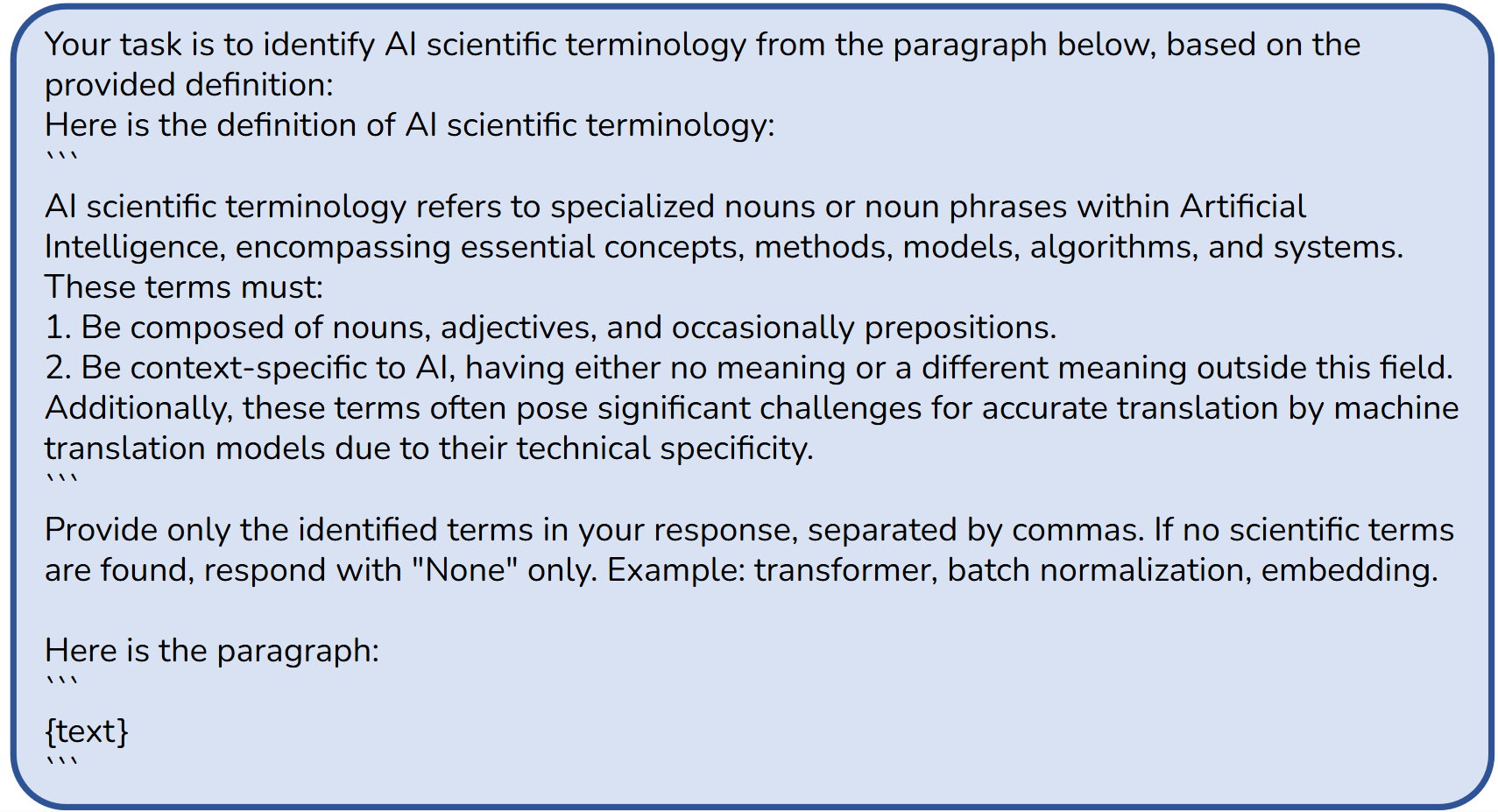}
    \caption{Prompts for extracting AI terminologies with \texttt{LLaMA-3-70B-Instruct}.}
    \label{fig:term_extraction_prompt}
\end{figure*}

\subsection{Terminology Translation Details}
\label{appn:term_translation_details}

\paragraph{Terminology Dataset Collection} In this section, we provide a detailed description of our methodology for creating the multilingual AI terminology dictionary and the associated translation experiments. The terms were extracted from papers published in the top AI conferences as shown in \cref{tab:top_ai_conference} across various fields including Artificial Intelligence, Computer Vision, Machine Learning, Natural Language Processing, and Web \& Information Retrieval. These conferences represent the leading venues in their respective domains.

\begin{table}[h]
\centering
\resizebox{\columnwidth}{!}{
\begin{tabular}{ll}
\toprule
Category & Selected Conferences \\
\midrule
AI & AAAI, IJCAI \\
CV & CVPR, ECCV, ICCV \\
ML & ICLR, ICML, NeurIPS, KDD \\
NLP & ACL, EMNLP, NAACL, EACL \\
& LREC, COLING, CoNLL \\
Web \& IR & SIGIR, WWW \\
\bottomrule
\end{tabular}
}
\caption{Top AI conferences included in our dataset collection.}
\label{tab:top_ai_conference}
\end{table}

\begin{figure*}[h]
    \centering
    \includegraphics[width=\linewidth]{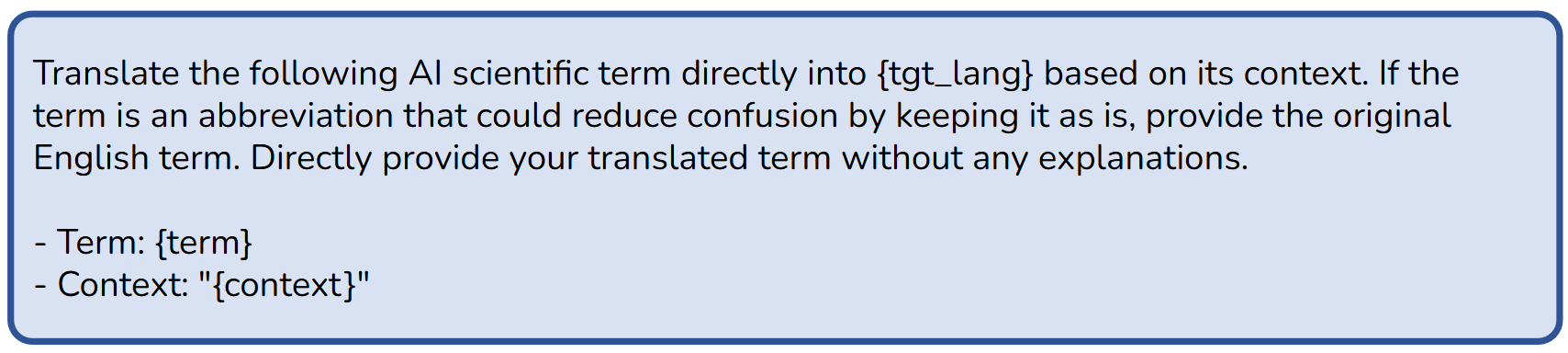}
    \caption{Prompts for translating AI terminologies with \texttt{GPT-3.5-Turbo} and \texttt{Claude 3 Sonnet}.}
    \label{fig:term_translation_prompt}
\end{figure*}

\paragraph{Prompt Design for Translation} 
To ensure consistent and high-quality translations, we used carefully designed prompts for GPT-3.5-Turbo and Claude 3 Sonnet. We provided sentence contexts relevant to each AI terminology, split into a maximum of three chunks, each containing up to 64 words. As shown in \cref{fig:term_translation_prompt}, the translation prompt asks the model to translate an AI-specific term into a target language, with context provided to clarify the meaning. If the term is an abbreviation or a technical term that should remain in its original form to avoid confusion, the models are instructed to retain the English term.

\begin{table*}[h]
\centering
\resizebox{0.8\textwidth}{!}{
\begin{tabular}{lrrrrr}
\toprule
 & Arabic & Chinese & French & Japanese & Russian \\
\midrule
Three-Model Agreement Ratio & 10.11\% & 42.71\% & 9.86\% & 16.60\% & 17.93\% \\
Two-Model Agreement Ratio & 30.52\% & 36.82\% & 45.15\% & 40.44\% & 38.36\% \\
\bottomrule
\end{tabular}
}
\caption{Translation agreement ratios among three models (Claude 3 Sonnet, Google Translate API, and GPT-3.5-Turbo) for five target languages. The table shows the ratio of terms where all three models agree (\textit{Three-Model Agreement Ratio}) and the ratio where any two models agree (\textit{Two-Model Agreement Ratio}).}
\label{tab:mt_agreement_ratio}
\end{table*}

\paragraph{Translation Agreement Analysis} We evaluated the consistency of translations across three models: Claude 3 Sonnet, GPT-3.5-Turbo, and Google Translate API. \cref{tab:mt_agreement_ratio} summarizes the agreement ratios for five target languages: Arabic, Chinese, French, Japanese, and Russian. The results indicate significant variation in agreement ratios across languages, with Chinese achieving the highest three-model agreement (42.71\%) and Arabic having the lowest (10.11\%). These findings underscore the need for human involvement in AI terminology translation, as automatic translations often fall short due to the inherent limitations of current models.

To further assess translation quality, we also examined which model's outputs align most closely with human annotations by incorporating \texttt{GPT-4o}. \cref{tab:llm_translation_comparison} presents the results, showing that \texttt{GPT-4o} is the most advanced LLM available at the time of this study. This claim is further supported by relevant literature \citep{wang2024m4u, sato-etal-2024-tmu, zhang2024good, openai2024gpt4o}. However, our evaluation reinforces the argument that, despite GPT-4o’s advancements, it remains inadequate for precise terminology translation. This highlights the necessity of human annotations to ensure translation accuracy and reliability.

\begin{table*}[h]
    \centering
    \resizebox{\textwidth}{!}{
    \begin{tabular}{lcccc}
        \toprule
        Language & Claude 3 vs. Human & Google Translate vs. Human & GPT-3.5 vs. Human & \shortstack{GPT-4o vs. Human \\ (500 Random Examples)} \\
        \midrule
        Chinese   & 69.26\%  & 62.84\%  & 59.20\%  & 76.80\% \\
        Arabic    & 29.26\%  & 36.45\%  & 23.05\%  & 39.80\% \\
        French    & 57.06\%  & 51.44\%  & 14.17\%  & 58.20\% \\
        Japanese  & 57.59\%  & 49.65\%  & 34.07\%  & 67.80\% \\
        Russian   & 39.72\%  & 42.63\%  & 28.23\%  & 41.20\% \\
        \bottomrule
    \end{tabular}
    }
    \caption{Comparison of translation accuracy between Claude 3, Google Translate, GPT-3.5, and GPT-4o against human translations across different languages.}
    \label{tab:llm_translation_comparison}
\end{table*}

% The detailed discussion is in \cref{sec:term_translation}. todo

\begin{figure*}[h]
    \centering
    \includegraphics[width=\linewidth]{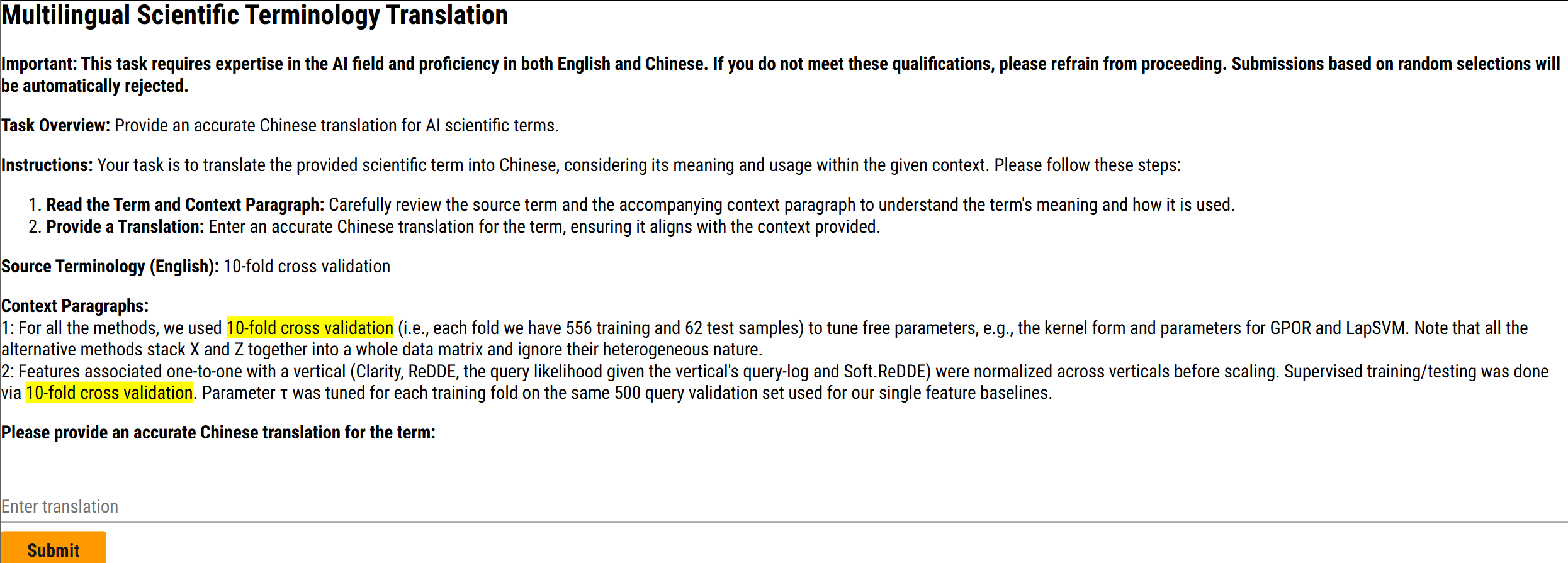}
    \caption{The MTurk layout demonstration for AI terminology translation generation task.}
    \label{fig:mturk_generation}
\end{figure*}

\paragraph{Human Translation via Mturk} To further validate the translations, we employed human translator through the MTurk platform. Each terminology was translated by 10 independent annotators, resulting in a set of 10 translation candidates for each term. The platform layout and guidance for helping translators give these translations is shown in \cref{fig:mturk_generation}. We compensate annotators in accordance with MTurk's payment standards.

\begin{figure*}[h]
    \centering
    \includegraphics[width=\linewidth]{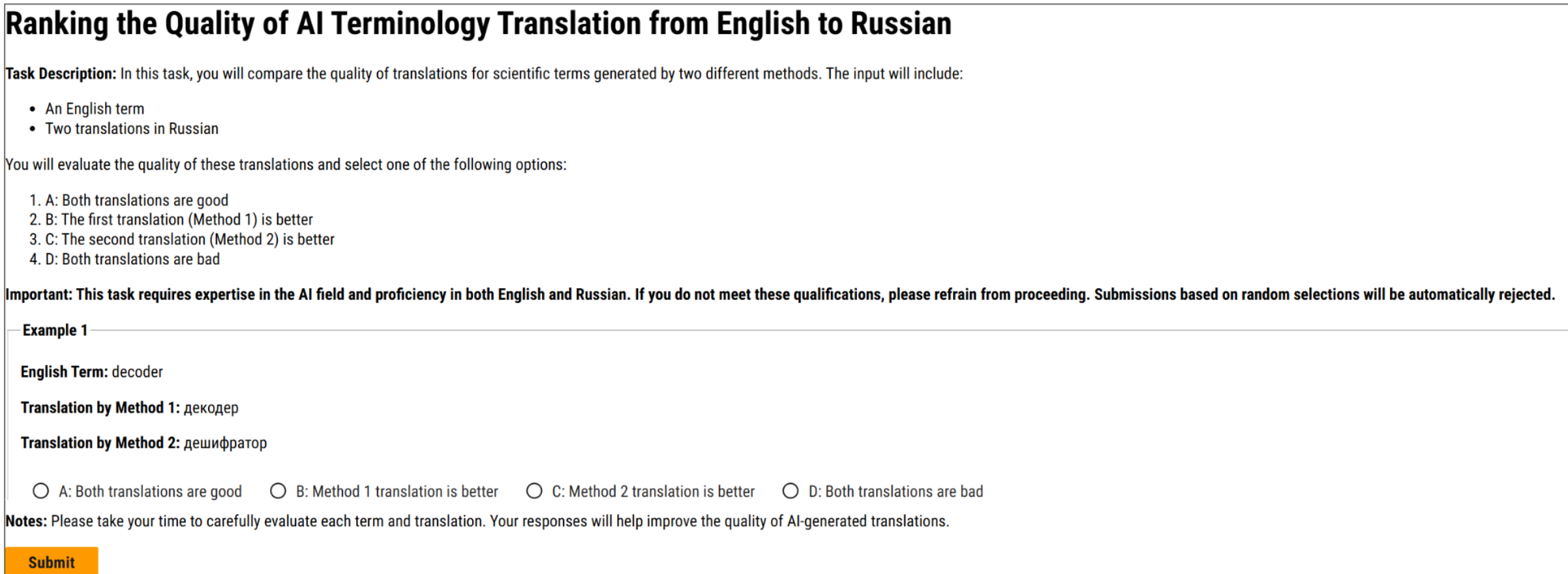}
    \caption{The MTurk layout demonstration for evaluating AI terminology translations. The layout is used for two tasks: (1) comparing the GPT-4o-selected candidate with the majority-voted candidate, and (2) comparing the translations in our dataset with those in the 60-60 initiative evaluation set.}
    \label{fig:mturk_rating}
\end{figure*}

\subsection{Dataset Quality Assessment Details}
\label{appn:dataset_quality_assessment_details}

To evaluate the quality of translations in our dataset, we conducted two tasks involving human annotators. Annotators were presented with pairs of translations generated by different methods and were tasked with evaluating their relative quality. The analysis of their ratings is summarized in Tables~\ref{tab:mturk_rating_distribution_gpt4o}, \ref{tab:mturk_rating_distribution_6060}, and \ref{tab:mturk_rating_agreement}. Below, we describe the findings in detail.

\paragraph{Task 1: Comparison Between GPT-4o-Selected Candidates and Majority-Voted Translations}
In Task 1, we constructed the evaluation dataset by first randomly sampling 1,000 terms per language that had received majority votes. We then filtered out terms where both translation strategies produced identical results, leaving approximately 200 terms per language for human evaluation. Specifically, this resulted in 323 Arabic terms, 185 Chinese terms, 180 French terms, 206 Japanese terms, and 230 Russian terms.

Annotators compared translations selected by GPT-4o (Method 1) with majority-voted translations generated by human annotators (Method 2). The choices were: A. Both translations are good; B. Method 1's translation is better; C. Method 2's translation is better; D. Both translations are bad.

\cref{tab:mturk_rating_distribution_gpt4o} summarizes the distribution of annotators' choices for Task 1. Across all languages, the majority of annotations fell into category A, where both translations were rated as good. Japanese showed the highest percentage of agreement in this category (56.99\%), followed by Russian (54.43\%) and Chinese (50.59\%). Cases where Method 1 translations were rated as better (category B) ranged from 24.37\% (Japanese) to 30.26\% (Russian). Similarly, cases where Method 2 translations were rated as better (category C) ranged from 13.04\% (Russian) to 20.37\% (Arabic). The lowest percentage of responses was observed in category D, where both translations were rated as bad, accounting for less than 5\% of responses across all languages.

\paragraph{Task 2: Comparison Between Dataset Translations and 60-60 Initiative Translations}
In Task 2, the number of overlapping terms between the 60-60 evaluation set and our dataset varies across languages for two main reasons. First, some terminology translations appear in certain languages but not others due to the integration of external data sources, as shown in \cref{tab:data_stats}. Second, some translations are identical in both our dataset and the 60-60 evaluation set; these were removed prior to human evaluation. As a result, the final analysis includes 162 Arabic terms, 106 Chinese terms, 77 French terms, 103 Japanese terms, and 88 Russian terms.

Annotators compared translations from our dataset (Method 1) with those in the 60-60 initiative evaluation set (Method 2). The distribution of choices is shown in Table~\ref{tab:mturk_rating_distribution_6060}. In this task, a higher percentage of annotators preferred translations from Method 2 (category B) for most languages, especially in Chinese (43.02\%), French (43.64\%), and Russian (45.00\%). However, Japanese translations from Method 1 had a significantly higher percentage in category A, with 57.28\% of annotators agreeing that both translations were good. Instances where both translations were rated as bad (category D) remained low across all languages, ranging from 2.60\% to 5.68\%.

\begin{table}[h]
\centering
\resizebox{\columnwidth}{!}{
\begin{tabular}{lccccc}
\toprule
 & Arabic & Chinese & French & Japanese & Russian \\
 \midrule
Task 1 & 0.30 & 0.22 & 0.37 & 0.21 & 0.39 \\
Task 2 & 0.22 & 0.41 & 0.50 & 0.39 & 0.41 \\
 \bottomrule
\end{tabular}
}
\caption{Fleiss' Kappa scores for inter-annotator agreement among 5 annotators across each question in Task 1 and Task 2.}
\label{tab:mturk_rating_agreement}
\end{table}

\paragraph{Inter-Annotator Agreement}

To measure inter-annotator agreement, we calculated Fleiss' Kappa scores for each language in both tasks. Table~\ref{tab:mturk_rating_agreement} reports these values. Fleiss' Kappa values between $0.20 \leq \kappa < 0.40$ indicate fair agreement, while values between $0.40 \leq \kappa < 0.60$ indicate moderate agreement. In Task 1, Kappa scores ranged from 0.21 (Japanese) to 0.39 (Russian), showing fair agreement across languages. In Task 2, Kappa scores improved, ranging from 0.39 (Japanese) to 0.50 (French), indicating moderate agreement for most languages.

\paragraph{Conclusion}

The results demonstrate that our dataset's translations are of high quality, with a majority of annotators rating them as good. While inter-annotator agreement was fair in Task 1, moderate agreement was observed in Task 2, highlighting the robustness of our dataset compared to established benchmarks.

\subsection{Dataset Coverage Assessment Details}
\label{appn:dataset_coverage_assessment_details}

\begin{figure}[t]
    \centering
    \includegraphics[width=\columnwidth]{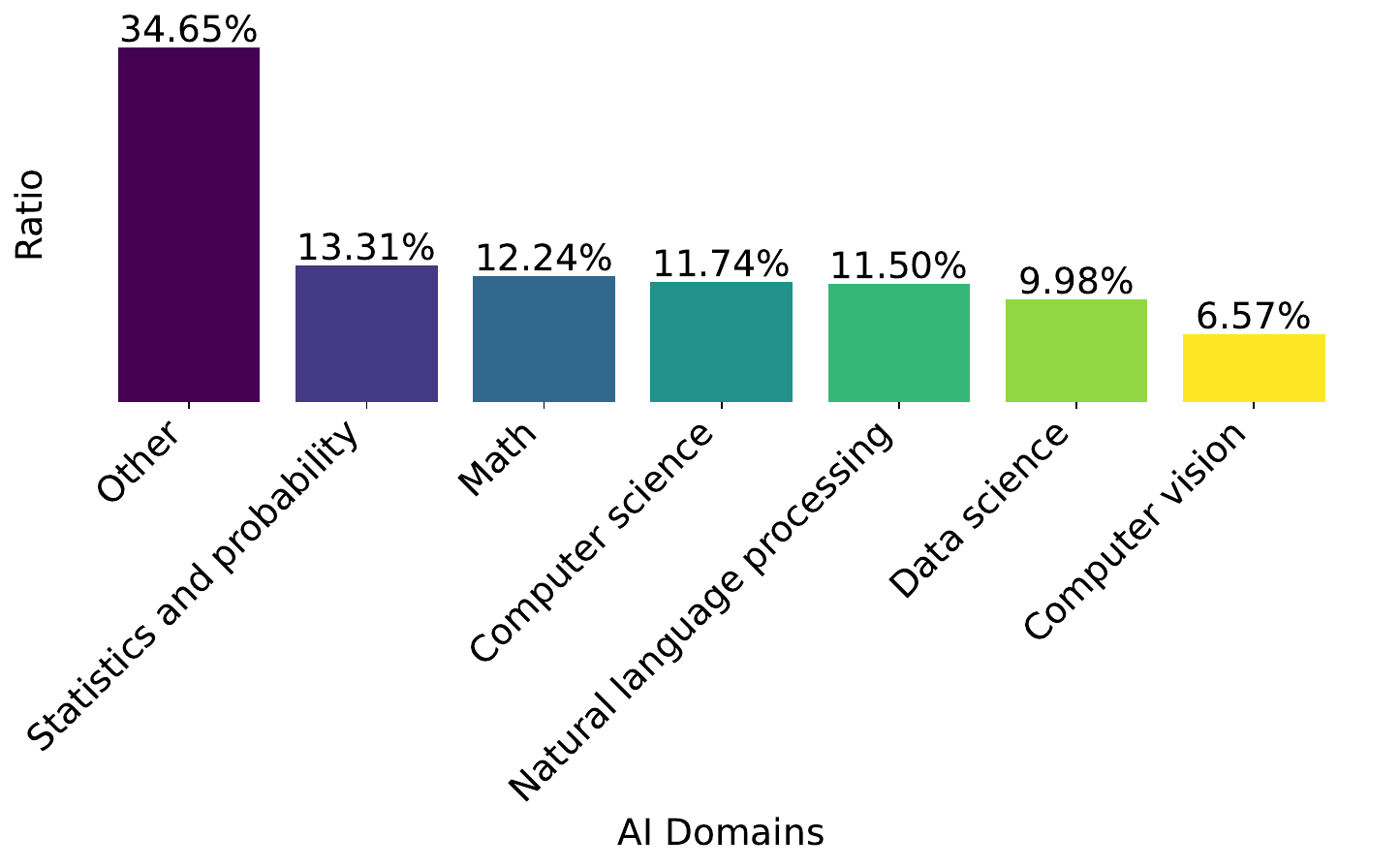}
    \caption{Terminology distribution of the top-6 AI domains in \ourdataset{}.}
    \label{fig:domain_coverage}
\end{figure}

\paragraph{Domain Distribution.}
The terminology in the dataset spans various AI-related domains. As shown in Figure~\ref{fig:domain_coverage}, the six most frequent domains include Statistics and Probability (13.31\%), Math (12.24\%), Computer Science (11.74\%), Natural Language Processing (11.50\%), Data Science (9.98\%), and Computer Vision (6.57\%). The largest proportion of terms (34.65\%) falls under the ``Other'' category, which represents interdisciplinary or less-defined concepts that do not fit neatly into any of the predefined categories. This distribution reflects the diversity and multidisciplinary nature of \ourdataset{}.

\begin{figure}[t]
    \centering
    \includegraphics[width=\columnwidth]{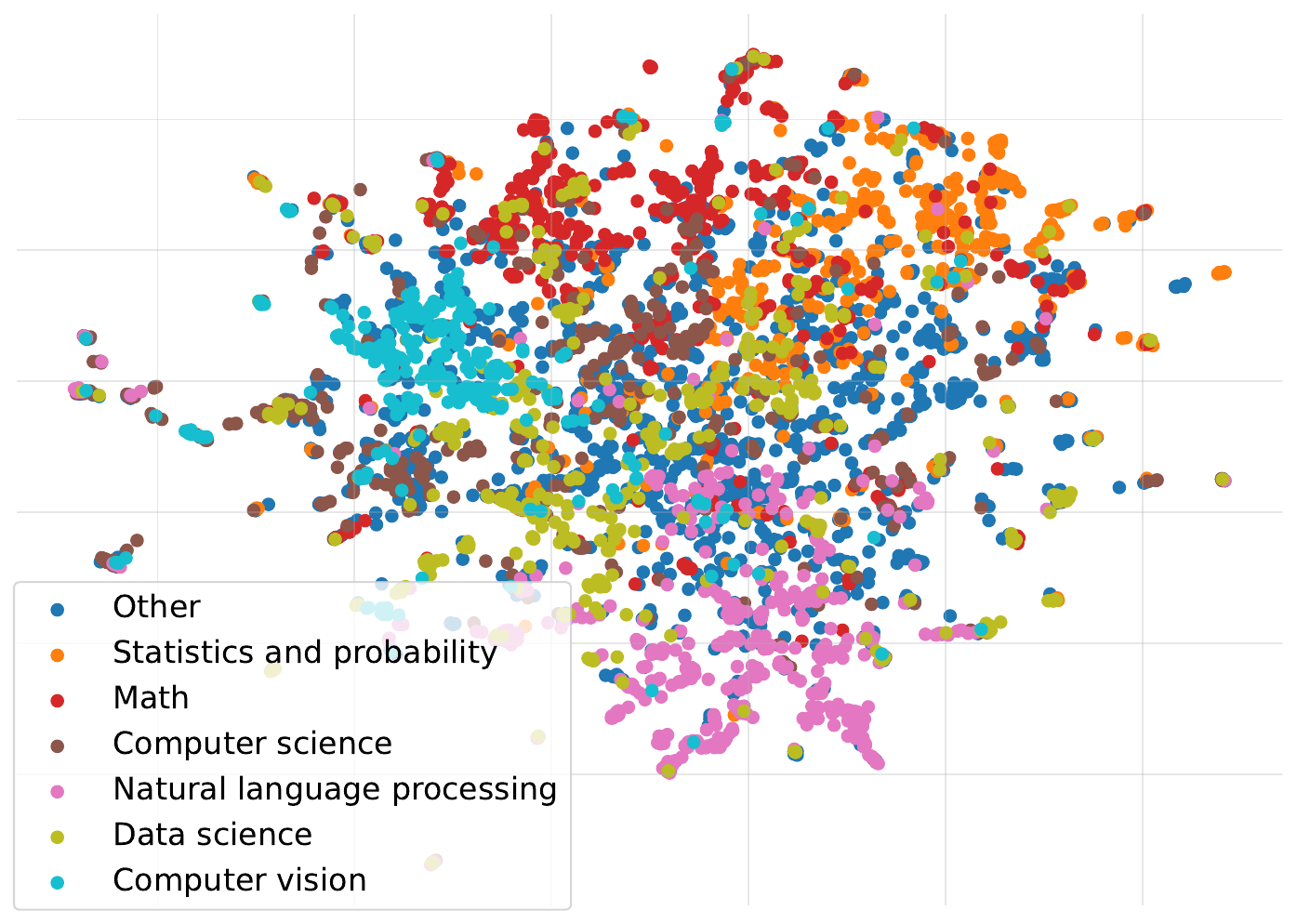}
    \caption{UMap visualization of terms in \ourdataset{} by domain.}
    \label{fig:domain_umap}
\end{figure}

\paragraph{Semantic Clustering.}
We used Uniform Manifold Approximation and Projection (UMAP) to visualize the semantic relationships between terms across domains. Figure~\ref{fig:domain_umap} shows a low-dimensional embedding of terms, where each point represents a term, color-coded by its domain. The visualization reveals distinct clusters corresponding to each domain, indicating strong intra-domain coherence. Overlaps between clusters (e.g., Data Science and Natural Language Processing) highlight the interconnected nature of these fields, where concepts are often shared or applied across domains.

\begin{figure}[t]
    \centering
    \includegraphics[width=\columnwidth]{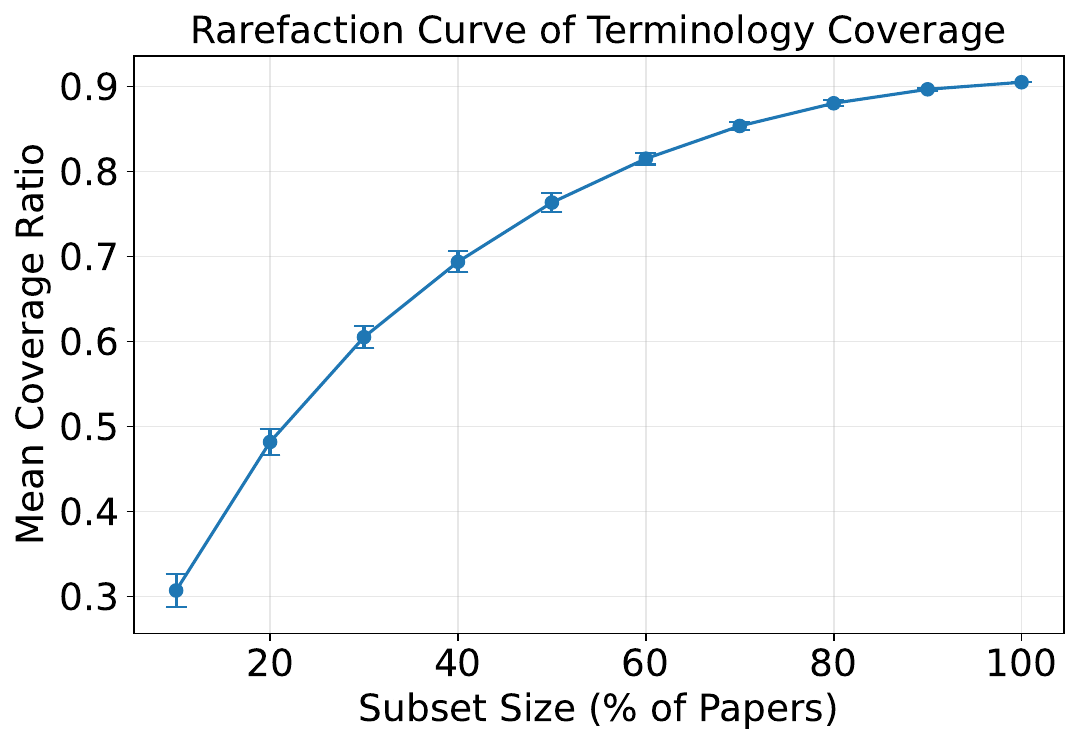}
    \caption{Rarefaction curve showing the mean coverage ratio of AI terminology with respect to the subset size of papers. Each subset size is sampled 50 times, and the error bars represent the standard deviation of the coverage ratios across these samples.}
    \label{fig:rarefaction_curve}
\end{figure}

\paragraph{Coverage Analysis through Rarefaction.}
To assess the comprehensiveness of our terminology collection, we conducted a rarefaction analysis, which is visualized in Figure~\ref{fig:rarefaction_curve}. This analysis reveals the relationship between the sample size of papers and the coverage of AI terminology. The curve demonstrates a characteristic asymptotic behavior, starting with a steep increase in coverage ratio from approximately 0.3 at 10\% of papers to 0.6 at 30\% of papers. As the subset size increases, the marginal gain in coverage gradually diminishes, reaching a coverage ratio of approximately 0.9 at 100\% of the papers. The error bars, representing standard deviation across 50 random samples for each subset size, notably decrease as the sample size increases, indicating more stable coverage at larger sample sizes. The asymptotic nature of the curve approaching 0.9 coverage suggests that our current collection has achieved a robust representation of commonly used AI terminology, with additional papers likely to introduce increasingly specialized or niche terms at a decreasing rate. The relatively small error bars at larger sample sizes (80-100\%) indicate high consistency in terminology coverage across different subsets of the literature, supporting the reliability of our collection methodology.

\section{Additional Experiment Details}

\subsection{Terminology Refinement via Prompting Details}
\label{appn:term_refinement_prompting_details}

\begin{figure*}[h]
    \centering
    \includegraphics[width=\linewidth]{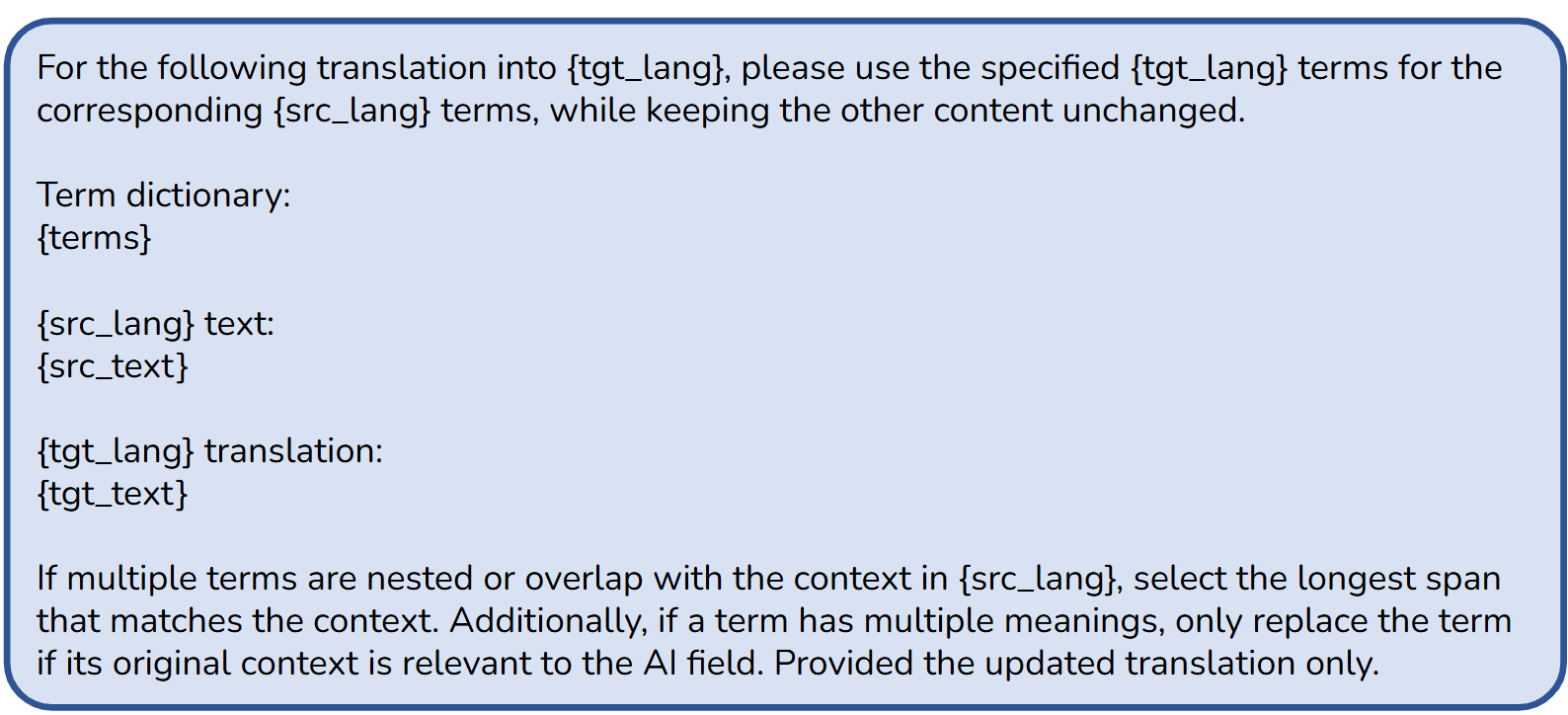}
    \caption{Prompts for refining a machine translation model's initial output using relevant term translations from \ourdataset{} as input.}
    \label{fig:term_refinement_prompt}
\end{figure*}

In this subsection, we describe the process of refining machine translation outputs by leveraging prompts tailored for terminology consistency. The objective is to ensure that specific domain-related terms are translated accurately, adhering to predefined mappings provided in the dataset.

As illustrated in \cref{fig:term_refinement_prompt}, the prompt incorporates a term dictionary, the source language \texttt{src\_lang} text, and the initial machine-translated \texttt{tgt\_lang} output. The model is instructed to revise the translation by applying the specified target terms for corresponding source terms while maintaining the rest of the content unchanged. This approach ensures consistency and accuracy in translation outputs, particularly in specialized fields, by integrating domain knowledge directly into the model's refinement process.

\subsection{Terminology Substitution by Word Alignment Details}
\label{appn:term_word_alignment_details}

\begin{figure*}[h]
    \centering
    \includegraphics[width=\linewidth]{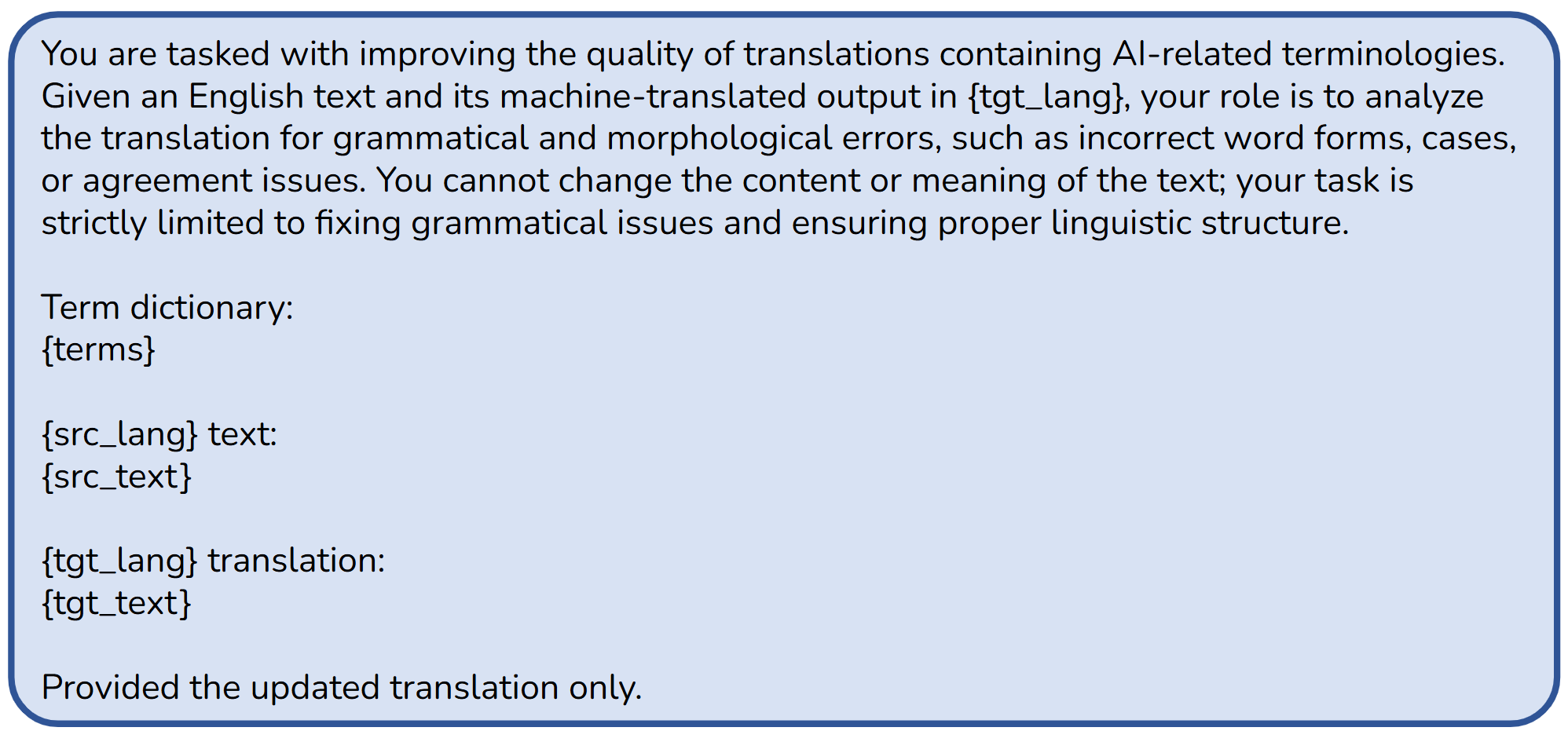}
    \caption{Prompts for refining the output of the word alignment method to ensure morphological coherence and accuracy.}
    \label{fig:word_alignment_prompt}
\end{figure*}

To validate our analysis in \cref{subsec:exp_results} regarding the varying performance of the word alignment method across different languages, we perform a post-hoc prompting step after word alignment. This step ensures that the translations are morphologically coherent and accurate, and is used to compare with the original results. The prompt is depicted in \cref{fig:word_alignment_prompt}. Refer to \cref{appn:additional_quantitative_res} for complete experiment results.

\subsection{Terminology-Aware Decoding Approaches}
\label{appn:term_decoding}

For constrained beam search \citep{hokamp-liu-2017-lexically}, we use the Hugging Face implementation, which enforces the inclusion of term translations from \ourdataset{} in the output.\footnote{\url{https://huggingface.co/blog/constrained-beam-search}} For token-level adjustment, we first identify AI terminology and their corresponding expected translations from \ourdataset{}. We then modify the output logits of these tokens by increasing them by factors of 10/7, 10/8, and 10/9 to prioritize their selection during decoding. See \cref{appn:additional_qualitative_res} for a discussion on performance.

\subsection{Evaluation Set Creation Details}
\label{appn:evaluation_set_creation_details}
The AI Papers and Model Cards evaluation set is created using 50 held-out AI research papers and 50 model cards generated by \citet{liu-etal-2024-automatic}. We then segment the content into sentence chunks, ensuring that each chunk contains no more than 64 words. From this set, we randomly sample 500 sentence chunks for evaluation.

% todo: add gpt-4o, also in the main text
% reviewer x1y4 Q3, vpaV Q2

For both evaluation sets, we conducted a manual evaluation, involving a new group of five expert annotators on MTurk. We randomly sampled 50 examples from the 60-60 test set and another 50 examples from the AI papers + model card test set. The annotators were provided with the English sentences, their corresponding translations, the associated terminologies, and their translations. The evaluation included the following tasks:

\begin{itemize}
    \item Task 1: Overall Translation Quality (rated on a scale from 1 to 5)
    \begin{itemize}
        \item 1 = The translation makes no sense
        \item 2 = Poor translation quality
        \item 3 = Acceptable translation quality
        \item 4 = Good translation quality
        \item 5 = Excellent translation quality
    \end{itemize}
    
    \item Task 2: Grammatical and Morphological Correctness (binary annotation: Yes/No)
    
    \item Task 3: Accuracy of Terminology Translations (binary annotation: Yes/No)
\end{itemize}

The results of the manual evaluation are summarized in \cref{tab:manual_eval}. We observe consistently high performance across all languages in terms of translation quality, grammatical correctness, and terminology accuracy, highlighting the robustness and linguistic reliability of ground truth labels generated in our evaluation set. We don't measure the impact of incorporating \ourdataset{} because there are only two possible reasons why translation quality might decrease after using GPT-4o to integrate terminologies: (1) low-quality terminology translations in our glossary, and (2) GPT-4o failing to integrate terminologies while maintaining grammatical accuracy. Regarding (1), we have already validated our glossary’s translation quality in \cref{sec:data_quality}. Regarding (2), terminology integration is a straightforward task, and it is unlikely that GPT-4o would significantly degrade grammatical accuracy.

\begin{table*}[h]
    \centering
    \begin{tabular}{lccc}
        \toprule
        \textbf{Language} & \textbf{Task 1 (Mean $\pm$ Std)} & \textbf{Task 2 (Accuracy)} & \textbf{Task 3 (Accuracy)} \\
        \midrule
        Chinese    & 3.96 $\pm$ 1.03 & 0.88 $\pm$ 0.32 & 0.91 $\pm$ 0.29 \\
        Arabic     & 3.39 $\pm$ 1.27 & 0.90 $\pm$ 0.30 & 0.91 $\pm$ 0.28 \\
        French     & 3.81 $\pm$ 1.20 & 0.78 $\pm$ 0.42 & 0.80 $\pm$ 0.40 \\
        Japanese   & 3.24 $\pm$ 1.13 & 0.77 $\pm$ 0.42 & 0.72 $\pm$ 0.45 \\
        Russian    & 3.11 $\pm$ 1.55 & 0.83 $\pm$ 0.38 & 0.82 $\pm$ 0.39 \\
        \bottomrule
    \end{tabular}
    \caption{Manual evaluation results for translation quality, grammatical correctness, and terminology accuracy across five languages.}
    \label{tab:manual_eval}
\end{table*}

\section{Additional Experiment Results}
\label{appn:additional_exp_results}

% We present comprehensive evaluation results using BLEU \citep{papineni-etal-2002-bleu}, ChrF \citep{popovic-2015-chrf}, ChrF++ \citep{popovic-2017-chrf}, TER \citep{snover-etal-2006-study}, and COMET \citep{rei2020comet} for translation into the five target languages across both evaluation sets from \cref{tab:bleu_full_arabic_nllb} to \cref{tab:comet_ai_papers}. We also present one good example and one bad example for the prompting-powered refinement and the word alignment method across all languages, as shown from \cref{fig:example_arabic} to \cref{fig:example_russian}.

\subsection{Additional Quantitative Results}
\label{appn:additional_quantitative_res}

We present comprehensive evaluation results using ChrF \citep{popovic-2015-chrf}, ChrF++ \citep{popovic-2017-chrf}, and TER \citep{snover-etal-2006-study} for translation into the five target languages across both evaluation sets in \cref{tab:exp_results_full}. By comparing the fourth and second values in each column of \cref{tab:exp_results}, we observe that the post-hoc prompting ablation enhances translation scores, aligning them with the performance of the prompting-powered method across all languages and models.

\begin{figure*}[t]
    \centering
    \includegraphics[width=\linewidth]{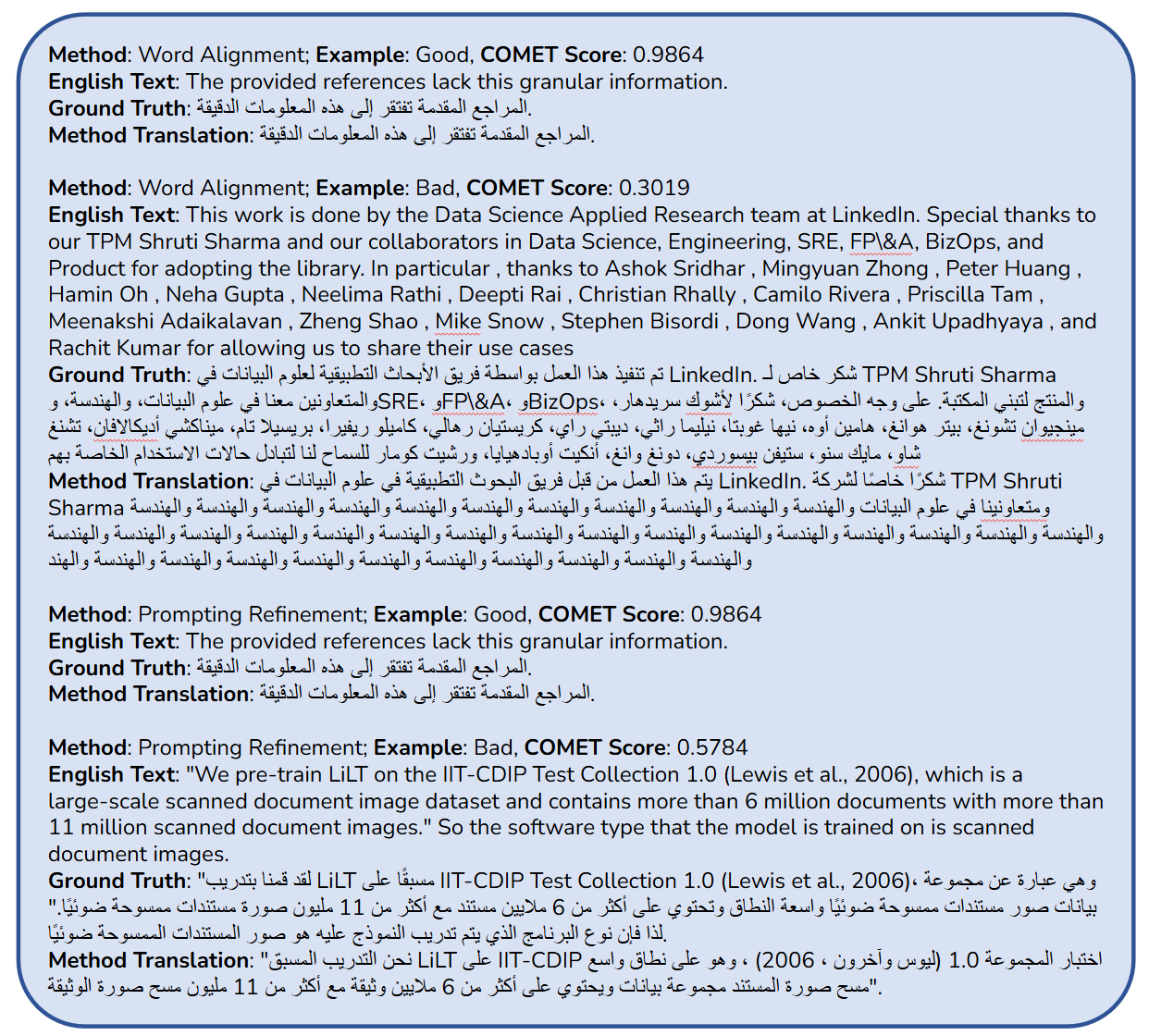}
    \caption{Arabic translation examples of the two integration cases on \texttt{nllb}, including one good example and one bad example for each case.}
    \label{fig:example_arabic}
\end{figure*}

\begin{figure*}[t]
    \centering
    \includegraphics[width=\linewidth]{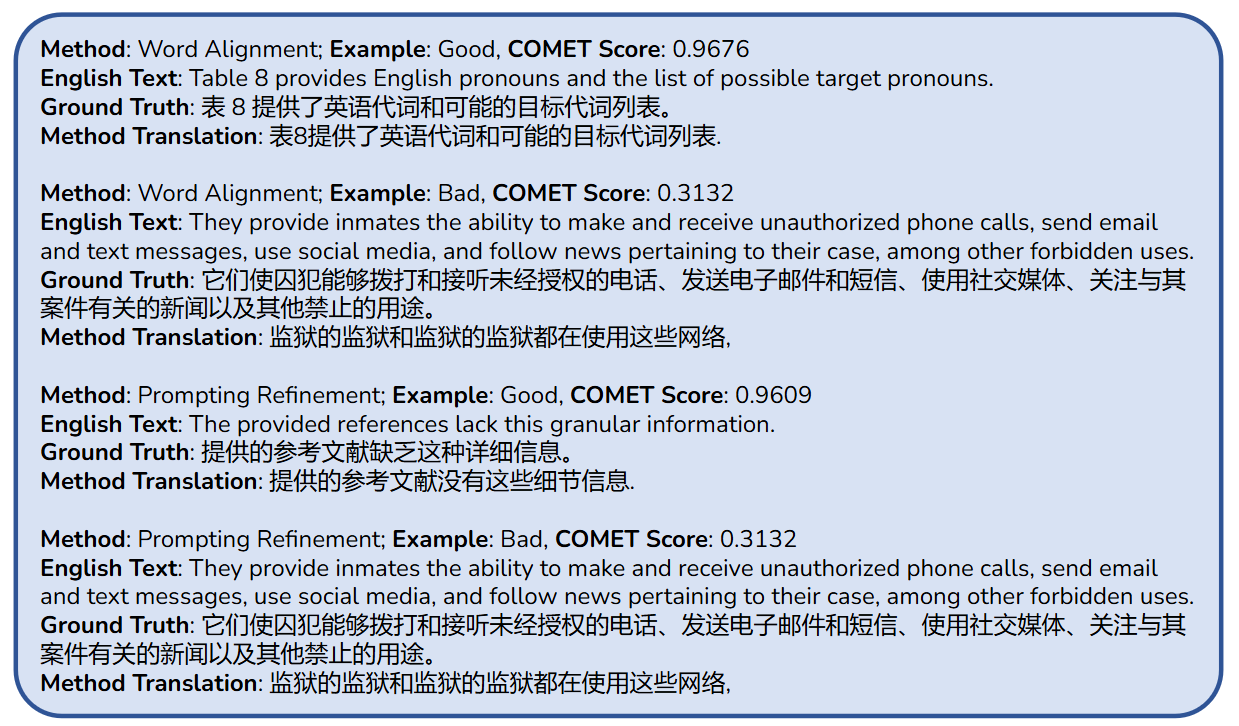}
    \caption{Chinese translation examples of the two integration cases on \texttt{nllb}, including one good example and one bad example for each case.}
    \label{fig:example_chinese}
\end{figure*}

\begin{figure*}[t]
    \centering
    \includegraphics[width=\linewidth]{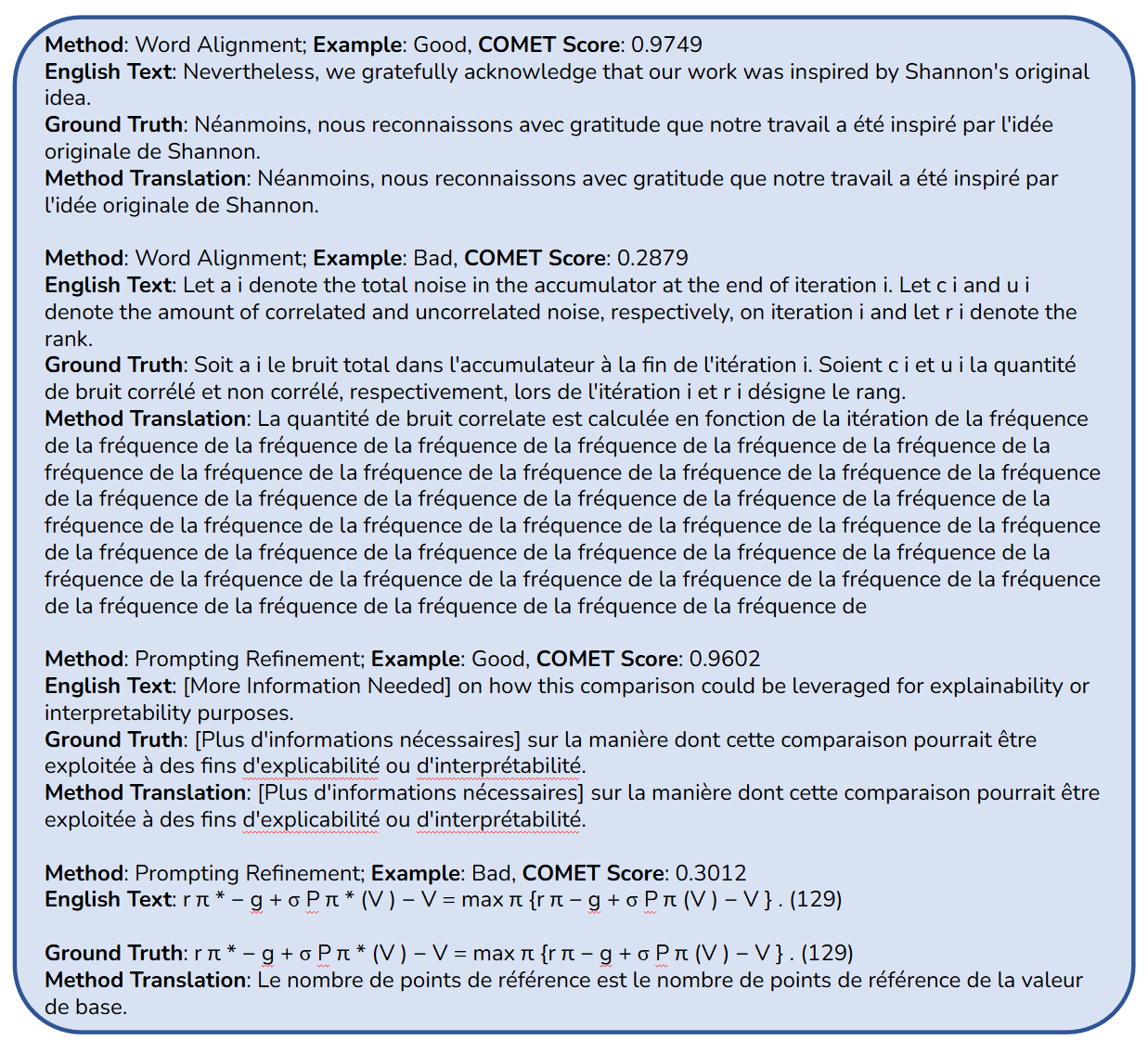}
    \caption{French translation examples of the two integration cases on \texttt{nllb}, including one good example and one bad example for each case.}
    \label{fig:example_french}
\end{figure*}

\begin{figure*}[t]
    \centering
    \includegraphics[width=\linewidth]{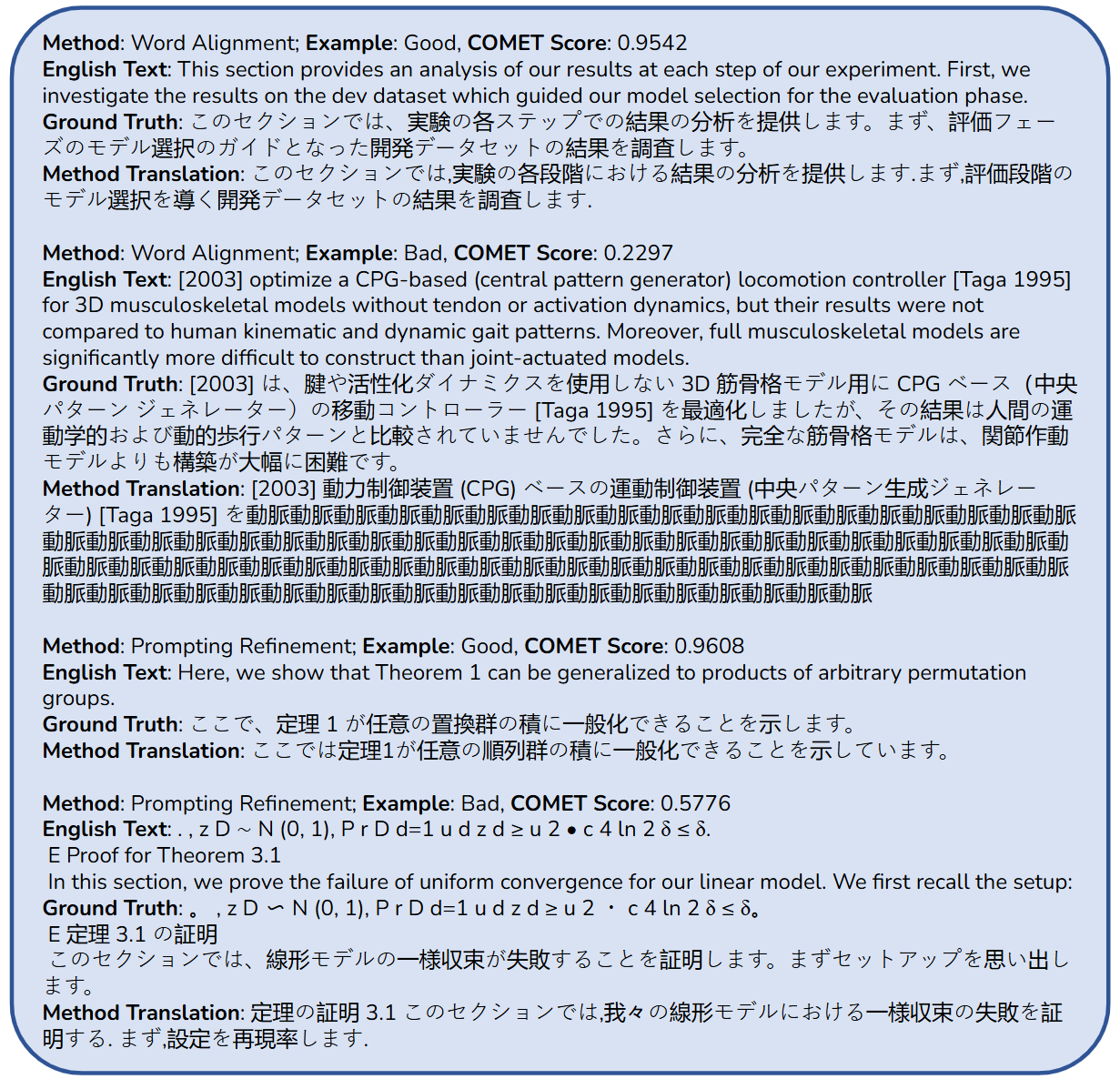}
    \caption{Japanese translation examples of the two integration cases on \texttt{nllb}, including one good example and one bad example for each case.}
    \label{fig:example_japanese}
\end{figure*}

\begin{figure*}[t]
    \centering
    \includegraphics[width=\linewidth]{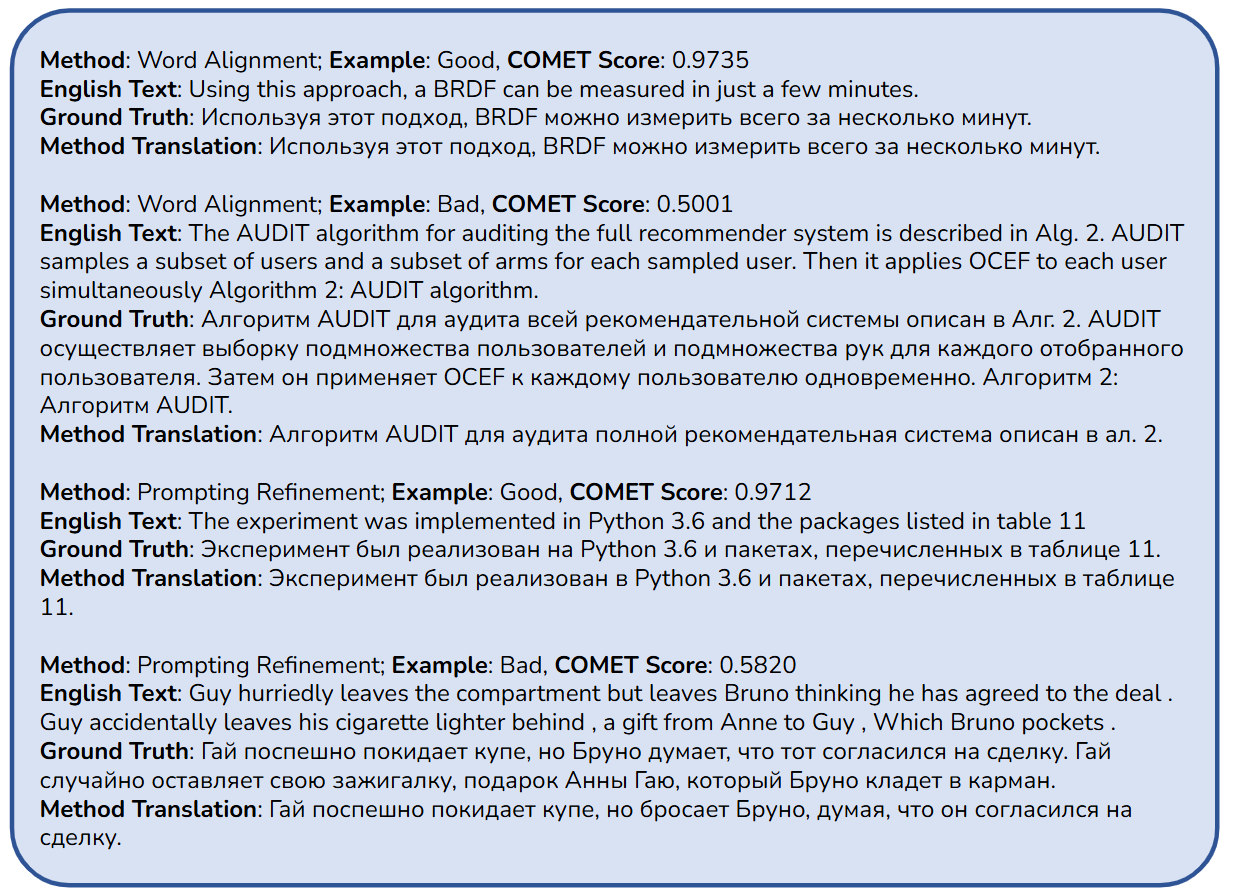}
    \caption{Russian translation examples of the two integration cases on \texttt{nllb}, including one good example and one bad example for each case.}
    \label{fig:example_russian}
\end{figure*}

We also present one good example and one bad example for the prompting-powered refinement and the word alignment method across all languages, as shown from \cref{fig:example_arabic} to \cref{fig:example_russian}.

\subsection{Qualitative Results}
\label{appn:additional_qualitative_res}

We also conduct a qualitative manual evaluation of the generation results produced by constrained beam search and logit adjustment methods, as explained in \cref{appn:term_decoding}. Both approaches exhibit extremely slow performance, running approximately 100 times slower than post-hoc methods. Furthermore, forcing specific word ids to appear in the output did not perform well, suffering from similar issues as the word alignment method in disrupting syntactical dependencies and morphological agreements. During manual inspection, we observe that constrained beam search behaves similarly to the logit adjustment method with a large scaling factor: the generated sentences often failed to maintain proper grammar, and terminological terms were frequently repeated multiple times within the same output. Conversely, with a small scaling factor in the logit adjustment method, the terms were often omitted entirely, demonstrating little to no effect on the output.

\subsection{Statistical Tests for Experiment Results}
\label{appn:stat_test_res}

To assess the statistical significance of our findings in \cref{subsec:exp_results}, we conducted one-sided paired t-tests on the BLEU, COMET, ChrF, ChrF++, and TER scores. Based on our observations, we formulated the following three hypotheses:

\begin{enumerate}[left=-2pt, label=$H_0^{(\arabic*)}$:]
    \item The prompting-powered refinement method outperforms direct translation across all languages.
    \item The word alignment method outperforms direct translation for Chinese and Japanese but underperforms for Arabic, French, and Russian.
    \item The prompting-powered refinement method outperforms the word alignment method across all languages.

\end{enumerate}

As shown in \cref{tab:hypothesis_test_stats}, hypotheses 1 and 3 are fully supported by all metrics across all languages. Hypothesis 2 is partially supported, as there are cases where we cannot reject the null hypothesis that the word alignment method underperforms default translation for Arabic, French, and Russian. This outcome is insightful, as it suggests that the default translation approach remains effective for these languages.

\begin{table*}[h]
\centering
\resizebox{\textwidth}{!}{
\begin{tabular}{cccccccccccc}
\toprule
& & \multicolumn{2}{c}{Arabic} & \multicolumn{2}{c}{Chinese} & \multicolumn{2}{c}{French} & \multicolumn{2}{c}{Japanese} & \multicolumn{2}{c}{Russian} \\
\midrule
Metric & Hypothesis & T Stats & P Val & T Stats & P Val & T Stats & P Val & T Stats & P Val & T Stats & P Val \\
\midrule
\multirow{3}{*}{COMET} & 1 & 14.18 & 0.00 & 15.22 & 0.00 & 12.68 & 0.00 & 11.13 & 0.00 & 8.53 & 0.00 \\
& 2 & 6.91 & 0.00 & 7.60 & 0.00 & 1.84 & 0.03 & 3.21 & 0.00 & 20.76 & 0.00 \\
& 3 & 17.93 & 0.00 & 11.67 & 0.00 & 13.58 & 0.00 & 10.62 & 0.00 & 24.89 & 0.00 \\
\midrule
\multirow{3}{*}{BLEU} & 1 & 6.37 & 0.00 & 6.72 & 0.00 & 9.80 & 0.00 & 6.26 & 0.00 & 7.63 & 0.00 \\
& 2 & -0.66 & 0.75 & 3.89 & 0.00 & -0.22 & 0.59 & 2.58 & 0.00 & -2.85 & 1.00 \\
& 3 & 4.80 & 0.00 & 3.87 & 0.00 & 8.77 & 0.00 & 4.57 & 0.00 & 5.50 & 0.00 \\

\midrule
\multirow{3}{*}{ChrF} & 1 & 6.37 & 0.00 & 6.72 & 0.00 & 9.79 & 0.00 & 6.26 & 0.00 & 7.64 & 0.00 \\
& 2 & -0.66 & 0.75 & 3.89 & 0.00 & -0.18 & 0.57 & 2.59 & 0.00 & -2.85 & 1.00 \\
& 3 & 4.79 & 0.00 & 3.86 & 0.00 & 8.76 & 0.00 & 4.55 & 0.00 & 5.49 & 0.00 \\

\midrule
\multirow{3}{*}{ChrF++} & 1 & 6.37 & 0.00 & 6.72 & 0.00 & 9.79 & 0.00 & 6.26 & 0.00 & 7.64 & 0.00 \\
& 2 & -0.66 & 0.75 & 3.89 & 0.00 & -0.18 & 0.57 & 2.59 & 0.00 & -2.85 & 1.00 \\
& 3 & 4.79 & 0.00 & 3.86 & 0.00 & 8.76 & 0.00 & 4.55 & 0.00 & 5.49 & 0.00 \\

\midrule
\multirow{3}{*}{TER} & 1 & 6.17 & 0.00 & 6.72 & 0.00 & 9.71 & 0.00 & 6.26 & 0.00 & 7.41 & 0.00 \\
& 2 & -0.85 & 0.80 & 3.88 & 0.00 & -0.18 & 0.57 & 2.59 & 0.00 & -2.76 & 1.00 \\
& 3 & 4.40 & 0.00 & 3.87 & 0.00 & 8.73 & 0.00 & 4.56 & 0.00 & 5.27 & 0.00 \\

\bottomrule
\end{tabular}
}
\caption{Hypothesis test statistics and p-values for all metrics across the five tested models.}
\label{tab:hypothesis_test_stats}
\end{table*}

\clearpage

% Full table 2

\begin{table*}[h]
\centering
\resizebox{\linewidth}{!}{

\begin{tabular}{llccccc}
\toprule
Model & Metric & Arabic & Chinese & French & Japanese & Russian \\

\midrule

\multicolumn{7}{l}{Evaluation Set: 60-60} \\
\specialrule{1.2pt}{0pt}{2.5pt}

\multirow{5}{*}{\texttt{aya-expanse}} & BLEU & 20.11 \textcolor{ForestGreen}{+ 1.23} \textcolor{ForestGreen}{+ 0.18} \textcolor{ForestGreen}{+ 1.24}  & 27.31 \textcolor{ForestGreen}{+ 1.33} \textcolor{ForestGreen}{+ 0.24} \textcolor{ForestGreen}{+ 0.53}  & 33.05 \textcolor{ForestGreen}{+ 2.46} \textcolor{ForestGreen}{+ 0.20} \textcolor{ForestGreen}{+ 1.97}  & 14.59 \textcolor{ForestGreen}{+ 0.61} \textcolor{ForestGreen}{+ 0.32} \textcolor{ForestGreen}{+ 1.08}  & 16.59 \textcolor{ForestGreen}{+ 1.59} \textcolor{red}{- 0.05} \textcolor{ForestGreen}{+ 0.92}  \\
& ChrF & 20.62 \textcolor{ForestGreen}{+ 1.24} \textcolor{ForestGreen}{+ 0.18} \textcolor{ForestGreen}{+ 1.26}  & 27.52 \textcolor{ForestGreen}{+ 1.32} \textcolor{ForestGreen}{+ 0.24} \textcolor{ForestGreen}{+ 0.52}  & 33.68 \textcolor{ForestGreen}{+ 2.44} \textcolor{ForestGreen}{+ 0.18} \textcolor{ForestGreen}{+ 1.96}  & 14.76 \textcolor{ForestGreen}{+ 0.61} \textcolor{ForestGreen}{+ 0.32} \textcolor{ForestGreen}{+ 1.08}  & 16.99 \textcolor{ForestGreen}{+ 1.59} \textcolor{red}{- 0.06} \textcolor{ForestGreen}{+ 0.92}  \\
& ChrF++ & 20.62 \textcolor{ForestGreen}{+ 1.24} \textcolor{ForestGreen}{+ 0.18} \textcolor{ForestGreen}{+ 1.26}  & 27.52 \textcolor{ForestGreen}{+ 1.32} \textcolor{ForestGreen}{+ 0.24} \textcolor{ForestGreen}{+ 0.52}  & 33.68 \textcolor{ForestGreen}{+ 2.44} \textcolor{ForestGreen}{+ 0.18} \textcolor{ForestGreen}{+ 1.96}  & 14.76 \textcolor{ForestGreen}{+ 0.61} \textcolor{ForestGreen}{+ 0.32} \textcolor{ForestGreen}{+ 1.08}  & 16.99 \textcolor{ForestGreen}{+ 1.59} \textcolor{red}{- 0.06} \textcolor{ForestGreen}{+ 0.92}  \\
& COMET & 81.96 \textcolor{ForestGreen}{+ 0.71} \textcolor{red}{- 0.52} \textcolor{ForestGreen}{+ 1.27}  & 83.43 \textcolor{ForestGreen}{+ 1.57} \textcolor{ForestGreen}{+ 0.08} \textcolor{ForestGreen}{+ 1.63}  & 81.83 \textcolor{ForestGreen}{+ 1.06} \textcolor{red}{- 0.11} \textcolor{ForestGreen}{+ 1.35}  & 88.54 \textcolor{ForestGreen}{+ 0.32} \textcolor{red}{- 0.01} \textcolor{ForestGreen}{+ 0.38}  & 82.27 \textcolor{ForestGreen}{+ 0.69} \textcolor{red}{- 2.02} \textcolor{ForestGreen}{+ 0.98}  \\
& TER & 93.61 \textcolor{ForestGreen}{- 1.35} \textcolor{ForestGreen}{- 0.24} \textcolor{ForestGreen}{- 1.36}  & 73.47 \textcolor{ForestGreen}{- 1.33} \textcolor{ForestGreen}{- 0.24} \textcolor{ForestGreen}{- 0.53}  & 77.43 \textcolor{ForestGreen}{- 2.78} \textcolor{ForestGreen}{- 0.23} \textcolor{ForestGreen}{- 2.28}  & 86.42 \textcolor{ForestGreen}{- 0.61} \textcolor{ForestGreen}{- 0.33} \textcolor{ForestGreen}{- 1.08}  & 94.56 \textcolor{ForestGreen}{- 1.74} \textcolor{red}{+ 0.08} \textcolor{ForestGreen}{- 0.96}  \\

\cline{2-7}

\multirow{5}{*}{\texttt{aya-23-8B}} &  BLEU & 19.98 \textcolor{ForestGreen}{+ 0.54} \textcolor{red}{- 0.21} \textcolor{ForestGreen}{+ 0.58}  & 26.08 \textcolor{ForestGreen}{+ 0.47} \textcolor{ForestGreen}{+ 0.39} \textcolor{ForestGreen}{+ 0.45}  & 33.85 \textcolor{ForestGreen}{+ 2.28} \textcolor{red}{- 0.11} \textcolor{ForestGreen}{+ 2.48}  & 15.06 \textcolor{ForestGreen}{+ 0.87} \textcolor{ForestGreen}{+ 0.36} \textcolor{ForestGreen}{+ 1.24}  & 15.77 \textcolor{ForestGreen}{+ 1.05} \textcolor{ForestGreen}{+ 0.37} \textcolor{ForestGreen}{+ 0.84}  \\
&  ChrF & 20.50 \textcolor{ForestGreen}{+ 0.54} \textcolor{red}{- 0.21} \textcolor{ForestGreen}{+ 0.58}  & 26.20 \textcolor{ForestGreen}{+ 0.47} \textcolor{ForestGreen}{+ 0.39} \textcolor{ForestGreen}{+ 0.45}  & 34.48 \textcolor{ForestGreen}{+ 2.26} \textcolor{red}{- 0.11} \textcolor{ForestGreen}{+ 2.46}  & 15.22 \textcolor{ForestGreen}{+ 0.87} \textcolor{ForestGreen}{+ 0.37} \textcolor{ForestGreen}{+ 1.24}  & 16.17 \textcolor{ForestGreen}{+ 1.06} \textcolor{ForestGreen}{+ 0.37} \textcolor{ForestGreen}{+ 0.85}  \\
&  ChrF++ & 20.50 \textcolor{ForestGreen}{+ 0.54} \textcolor{red}{- 0.21} \textcolor{ForestGreen}{+ 0.58}  & 26.20 \textcolor{ForestGreen}{+ 0.47} \textcolor{ForestGreen}{+ 0.39} \textcolor{ForestGreen}{+ 0.45}  & 34.48 \textcolor{ForestGreen}{+ 2.26} \textcolor{red}{- 0.11} \textcolor{ForestGreen}{+ 2.46}  & 15.22 \textcolor{ForestGreen}{+ 0.87} \textcolor{ForestGreen}{+ 0.37} \textcolor{ForestGreen}{+ 1.24}  & 16.17 \textcolor{ForestGreen}{+ 1.06} \textcolor{ForestGreen}{+ 0.37} \textcolor{ForestGreen}{+ 0.85}  \\
&  COMET & 84.02 \textcolor{ForestGreen}{+ 0.81} \textcolor{red}{- 0.24} \textcolor{ForestGreen}{+ 1.05}  & 85.12 \textcolor{ForestGreen}{+ 0.58} \textcolor{ForestGreen}{+ 0.38} \textcolor{ForestGreen}{+ 0.93}  & 82.40 \textcolor{ForestGreen}{+ 0.94} \textcolor{red}{- 0.15} \textcolor{ForestGreen}{+ 1.26}  & 87.92 \textcolor{ForestGreen}{+ 0.50} \textcolor{ForestGreen}{+ 0.09} \textcolor{ForestGreen}{+ 0.70}  & 81.91 \textcolor{ForestGreen}{+ 0.40} \textcolor{red}{- 2.26} \textcolor{ForestGreen}{+ 0.84}  \\
&  TER & 93.97 \textcolor{ForestGreen}{- 0.57} \textcolor{red}{+ 0.23} \textcolor{ForestGreen}{- 0.61}  & 74.40 \textcolor{ForestGreen}{- 0.46} \textcolor{ForestGreen}{- 0.38} \textcolor{ForestGreen}{- 0.44}  & 76.35 \textcolor{ForestGreen}{- 2.64} \textcolor{red}{+ 0.18} \textcolor{ForestGreen}{- 2.89}  & 85.73 \textcolor{ForestGreen}{- 0.88} \textcolor{ForestGreen}{- 0.37} \textcolor{ForestGreen}{- 1.24}  & 95.60 \textcolor{ForestGreen}{- 1.17} \textcolor{ForestGreen}{- 0.45} \textcolor{ForestGreen}{- 0.98}  \\

\cline{2-7}

\multirow{5}{*}{\texttt{gpt-4o-mini}} &  BLEU & 23.58 \textcolor{ForestGreen}{+ 1.07} \textcolor{red}{- 0.00} \textcolor{ForestGreen}{+ 0.50}  & 32.64 \textcolor{ForestGreen}{+ 1.60} \textcolor{ForestGreen}{+ 0.66} \textcolor{ForestGreen}{+ 1.48}  & 40.80 \textcolor{ForestGreen}{+ 3.08} \textcolor{ForestGreen}{+ 0.50} \textcolor{ForestGreen}{+ 2.38}  & 21.46 \textcolor{ForestGreen}{+ 0.64} \textcolor{ForestGreen}{+ 0.19} \textcolor{ForestGreen}{+ 0.94}  & 17.25 \textcolor{ForestGreen}{+ 1.07} \textcolor{red}{- 0.13} \textcolor{ForestGreen}{+ 0.56}  \\
&  ChrF & 24.06 \textcolor{ForestGreen}{+ 1.07} \textcolor{red}{- 0.00} \textcolor{ForestGreen}{+ 0.50}  & 32.76 \textcolor{ForestGreen}{+ 1.60} \textcolor{ForestGreen}{+ 0.66} \textcolor{ForestGreen}{+ 1.48}  & 41.40 \textcolor{ForestGreen}{+ 3.06} \textcolor{ForestGreen}{+ 0.49} \textcolor{ForestGreen}{+ 2.37}  & 21.59 \textcolor{ForestGreen}{+ 0.64} \textcolor{ForestGreen}{+ 0.19} \textcolor{ForestGreen}{+ 0.94}  & 17.65 \textcolor{ForestGreen}{+ 1.08} \textcolor{red}{- 0.13} \textcolor{ForestGreen}{+ 0.56}  \\
&  ChrF++ & 24.06 \textcolor{ForestGreen}{+ 1.07} \textcolor{red}{- 0.00} \textcolor{ForestGreen}{+ 0.50}  & 32.76 \textcolor{ForestGreen}{+ 1.60} \textcolor{ForestGreen}{+ 0.66} \textcolor{ForestGreen}{+ 1.48}  & 41.40 \textcolor{ForestGreen}{+ 3.06} \textcolor{ForestGreen}{+ 0.49} \textcolor{ForestGreen}{+ 2.37}  & 21.59 \textcolor{ForestGreen}{+ 0.64} \textcolor{ForestGreen}{+ 0.19} \textcolor{ForestGreen}{+ 0.94}  & 17.65 \textcolor{ForestGreen}{+ 1.08} \textcolor{red}{- 0.13} \textcolor{ForestGreen}{+ 0.56}  \\
&  COMET & 85.77 \textcolor{ForestGreen}{+ 0.69} \textcolor{red}{- 0.44} \textcolor{ForestGreen}{+ 0.61}  & 87.30 \textcolor{ForestGreen}{+ 0.48} \textcolor{ForestGreen}{+ 0.26} \textcolor{ForestGreen}{+ 0.44}  & 84.56 \textcolor{ForestGreen}{+ 0.68} \textcolor{red}{- 0.04} \textcolor{ForestGreen}{+ 0.62}  & 89.96 \textcolor{ForestGreen}{+ 0.14} \textcolor{ForestGreen}{+ 0.01} \textcolor{ForestGreen}{+ 0.14}  & 83.68 \textcolor{ForestGreen}{+ 0.38} \textcolor{red}{- 2.29} \textcolor{ForestGreen}{+ 0.47}  \\
&  TER & 89.66 \textcolor{ForestGreen}{- 1.17} \textcolor{ForestGreen}{- 0.11} \textcolor{ForestGreen}{- 0.54}  & 67.79 \textcolor{ForestGreen}{- 1.60} \textcolor{ForestGreen}{- 0.66} \textcolor{ForestGreen}{- 1.48}  & 68.32 \textcolor{ForestGreen}{- 3.57} \textcolor{ForestGreen}{- 0.56} \textcolor{ForestGreen}{- 2.76}  & 79.33 \textcolor{ForestGreen}{- 0.64} \textcolor{ForestGreen}{- 0.20} \textcolor{ForestGreen}{- 0.94}  & 93.90 \textcolor{ForestGreen}{- 1.17} \textcolor{red}{+ 0.15} \textcolor{ForestGreen}{- 0.58}  \\

\cline{2-7}

\multirow{5}{*}{\texttt{nllb}} & BLEU & 22.38 \textcolor{ForestGreen}{+ 1.37} \textcolor{ForestGreen}{+ 0.64} \textcolor{ForestGreen}{+ 1.21}  & 17.29 \textcolor{ForestGreen}{+ 1.92} \textcolor{ForestGreen}{+ 1.02} \textcolor{ForestGreen}{+ 2.40}  & 34.93 \textcolor{ForestGreen}{+ 2.86} \textcolor{ForestGreen}{+ 0.21} \textcolor{ForestGreen}{+ 3.23}  & 6.19 \textcolor{ForestGreen}{+ 2.42} \textcolor{ForestGreen}{+ 0.53} \textcolor{ForestGreen}{+ 2.95}  & 17.30 \textcolor{ForestGreen}{+ 1.54} \textcolor{ForestGreen}{+ 1.07} \textcolor{ForestGreen}{+ 1.51}  \\
& ChrF & 22.87 \textcolor{ForestGreen}{+ 1.37} \textcolor{ForestGreen}{+ 0.64} \textcolor{ForestGreen}{+ 1.22}  & 17.45 \textcolor{ForestGreen}{+ 1.92} \textcolor{ForestGreen}{+ 1.03} \textcolor{ForestGreen}{+ 2.40}  & 35.55 \textcolor{ForestGreen}{+ 2.85} \textcolor{ForestGreen}{+ 0.21} \textcolor{ForestGreen}{+ 3.21}  & 6.22 \textcolor{ForestGreen}{+ 2.42} \textcolor{ForestGreen}{+ 0.53} \textcolor{ForestGreen}{+ 2.94}  & 17.69 \textcolor{ForestGreen}{+ 1.55} \textcolor{ForestGreen}{+ 1.08} \textcolor{ForestGreen}{+ 1.53}  \\
& ChrF++ & 22.87 \textcolor{ForestGreen}{+ 1.37} \textcolor{ForestGreen}{+ 0.64} \textcolor{ForestGreen}{+ 1.22}  & 17.45 \textcolor{ForestGreen}{+ 1.92} \textcolor{ForestGreen}{+ 1.03} \textcolor{ForestGreen}{+ 2.40}  & 35.55 \textcolor{ForestGreen}{+ 2.85} \textcolor{ForestGreen}{+ 0.21} \textcolor{ForestGreen}{+ 3.21}  & 6.22 \textcolor{ForestGreen}{+ 2.42} \textcolor{ForestGreen}{+ 0.53} \textcolor{ForestGreen}{+ 2.94}  & 17.69 \textcolor{ForestGreen}{+ 1.55} \textcolor{ForestGreen}{+ 1.08} \textcolor{ForestGreen}{+ 1.53}  \\
& COMET & 83.52 \textcolor{ForestGreen}{+ 0.83} \textcolor{red}{- 0.45} \textcolor{ForestGreen}{+ 1.31}  & 78.22 \textcolor{ForestGreen}{+ 2.95} \textcolor{ForestGreen}{+ 0.73} \textcolor{ForestGreen}{+ 3.67}  & 82.83 \textcolor{ForestGreen}{+ 1.00} \textcolor{red}{- 0.19} \textcolor{ForestGreen}{+ 1.43}  & 77.82 \textcolor{ForestGreen}{+ 3.80} \textcolor{ForestGreen}{+ 0.39} \textcolor{ForestGreen}{+ 4.80}  & 81.41 \textcolor{ForestGreen}{+ 0.97} \textcolor{red}{- 1.54} \textcolor{ForestGreen}{+ 1.79}  \\
& TER & 91.12 \textcolor{ForestGreen}{- 1.59} \textcolor{ForestGreen}{- 0.74} \textcolor{ForestGreen}{- 1.34}  & 83.26 \textcolor{ForestGreen}{- 1.93} \textcolor{ForestGreen}{- 1.03} \textcolor{ForestGreen}{- 2.40}  & 75.21 \textcolor{ForestGreen}{- 3.24} \textcolor{ForestGreen}{- 0.26} \textcolor{ForestGreen}{- 3.72}  & 94.32 \textcolor{ForestGreen}{- 2.41} \textcolor{ForestGreen}{- 0.53} \textcolor{ForestGreen}{- 2.94}  & 93.87 \textcolor{ForestGreen}{- 1.67} \textcolor{ForestGreen}{- 1.17} \textcolor{ForestGreen}{- 1.65}  \\

\cline{2-7}

\multirow{5}{*}{\texttt{seamless}} & BLEU & 23.13 \textcolor{ForestGreen}{+ 1.16} \textcolor{red}{- 0.03} \textcolor{ForestGreen}{+ 2.35}  & 26.26 \textcolor{ForestGreen}{+ 0.97} \textcolor{ForestGreen}{+ 0.80} \textcolor{ForestGreen}{+ 2.75}  & 40.04 \textcolor{ForestGreen}{+ 2.08} \textcolor{red}{- 0.57} \textcolor{ForestGreen}{+ 1.69}  & 14.56 \textcolor{ForestGreen}{+ 0.74} \textcolor{ForestGreen}{+ 0.05} \textcolor{ForestGreen}{+ 2.38}  & 17.18 \textcolor{ForestGreen}{+ 1.71} \textcolor{ForestGreen}{+ 1.17} \textcolor{ForestGreen}{+ 1.22}  \\
& ChrF & 23.67 \textcolor{ForestGreen}{+ 1.17} \textcolor{red}{- 0.03} \textcolor{ForestGreen}{+ 2.36}  & 26.43 \textcolor{ForestGreen}{+ 0.98} \textcolor{ForestGreen}{+ 0.80} \textcolor{ForestGreen}{+ 2.74}  & 40.68 \textcolor{ForestGreen}{+ 2.08} \textcolor{red}{- 0.58} \textcolor{ForestGreen}{+ 1.67}  & 14.65 \textcolor{ForestGreen}{+ 0.74} \textcolor{ForestGreen}{+ 0.05} \textcolor{ForestGreen}{+ 2.37}  & 17.61 \textcolor{ForestGreen}{+ 1.71} \textcolor{ForestGreen}{+ 1.17} \textcolor{ForestGreen}{+ 1.23}  \\
& ChrF++ & 23.67 \textcolor{ForestGreen}{+ 1.17} \textcolor{red}{- 0.03} \textcolor{ForestGreen}{+ 2.36}  & 26.43 \textcolor{ForestGreen}{+ 0.98} \textcolor{ForestGreen}{+ 0.80} \textcolor{ForestGreen}{+ 2.74}  & 40.68 \textcolor{ForestGreen}{+ 2.08} \textcolor{red}{- 0.58} \textcolor{ForestGreen}{+ 1.67}  & 14.65 \textcolor{ForestGreen}{+ 0.74} \textcolor{ForestGreen}{+ 0.05} \textcolor{ForestGreen}{+ 2.37}  & 17.61 \textcolor{ForestGreen}{+ 1.71} \textcolor{ForestGreen}{+ 1.17} \textcolor{ForestGreen}{+ 1.23}  \\
& COMET & 84.07 \textcolor{ForestGreen}{+ 0.94} \textcolor{red}{- 0.38} \textcolor{ForestGreen}{+ 1.26}  & 83.44 \textcolor{ForestGreen}{+ 1.48} \textcolor{ForestGreen}{+ 0.50} \textcolor{ForestGreen}{+ 2.49}  & 83.86 \textcolor{ForestGreen}{+ 0.78} \textcolor{red}{- 0.07} \textcolor{ForestGreen}{+ 1.05}  & 85.05 \textcolor{ForestGreen}{+ 1.06} \textcolor{ForestGreen}{+ 0.16} \textcolor{ForestGreen}{+ 1.74}  & 82.33 \textcolor{ForestGreen}{+ 0.56} \textcolor{red}{- 1.87} \textcolor{ForestGreen}{+ 1.21}  \\
& TER & 90.27 \textcolor{ForestGreen}{- 1.33} \textcolor{red}{+ 0.01} \textcolor{ForestGreen}{- 2.78}  & 74.32 \textcolor{ForestGreen}{- 0.97} \textcolor{ForestGreen}{- 0.80} \textcolor{ForestGreen}{- 2.74}  & 69.19 \textcolor{ForestGreen}{- 2.38} \textcolor{red}{+ 0.65} \textcolor{ForestGreen}{- 1.91}  & 86.02 \textcolor{ForestGreen}{- 0.74} \textcolor{ForestGreen}{- 0.05} \textcolor{ForestGreen}{- 2.38}  & 94.01 \textcolor{ForestGreen}{- 1.89} \textcolor{ForestGreen}{- 1.30} \textcolor{ForestGreen}{- 1.38}  \\

\midrule

\multicolumn{7}{l}{Evaluation Set: AI Papers \& Model Cards} \\
\specialrule{1.2pt}{0pt}{2.5pt}

\multirow{5}{*}{\texttt{aya-expanse}} & BLEU & 11.47 \textcolor{ForestGreen}{+ 0.37} \textcolor{ForestGreen}{+ 0.10} \textcolor{ForestGreen}{+ 0.49}  & 12.04 \textcolor{ForestGreen}{+ 0.94} \textcolor{ForestGreen}{+ 0.17} \textcolor{ForestGreen}{+ 0.85}  & 18.84 \textcolor{ForestGreen}{+ 1.71} \textcolor{red}{- 0.85} \textcolor{ForestGreen}{+ 2.19}  & 8.11 \textcolor{red}{- 0.03} \textcolor{ForestGreen}{+ 0.04} \textcolor{ForestGreen}{+ 0.31}  & 13.84 \textcolor{ForestGreen}{+ 0.21} \textcolor{ForestGreen}{+ 0.32} \textcolor{ForestGreen}{+ 0.68}  \\
& ChrF & 11.95 \textcolor{ForestGreen}{+ 0.37} \textcolor{ForestGreen}{+ 0.10} \textcolor{ForestGreen}{+ 0.50}  & 12.76 \textcolor{ForestGreen}{+ 0.98} \textcolor{ForestGreen}{+ 0.18} \textcolor{ForestGreen}{+ 0.87}  & 19.49 \textcolor{ForestGreen}{+ 1.74} \textcolor{red}{- 0.85} \textcolor{ForestGreen}{+ 2.22}  & 8.59 \textcolor{red}{- 0.02} \textcolor{ForestGreen}{+ 0.06} \textcolor{ForestGreen}{+ 0.33}  & 14.32 \textcolor{ForestGreen}{+ 0.21} \textcolor{ForestGreen}{+ 0.34} \textcolor{ForestGreen}{+ 0.70}  \\
& ChrF++ & 11.95 \textcolor{ForestGreen}{+ 0.37} \textcolor{ForestGreen}{+ 0.10} \textcolor{ForestGreen}{+ 0.50}  & 12.76 \textcolor{ForestGreen}{+ 0.98} \textcolor{ForestGreen}{+ 0.18} \textcolor{ForestGreen}{+ 0.87}  & 19.49 \textcolor{ForestGreen}{+ 1.74} \textcolor{red}{- 0.85} \textcolor{ForestGreen}{+ 2.22}  & 8.59 \textcolor{red}{- 0.02} \textcolor{ForestGreen}{+ 0.06} \textcolor{ForestGreen}{+ 0.33}  & 14.32 \textcolor{ForestGreen}{+ 0.21} \textcolor{ForestGreen}{+ 0.34} \textcolor{ForestGreen}{+ 0.70}  \\
& COMET & 80.98 \textcolor{ForestGreen}{+ 0.46} \textcolor{red}{- 0.51} \textcolor{ForestGreen}{+ 0.63}  & 82.42 \textcolor{ForestGreen}{+ 0.56} \textcolor{red}{- 0.01} \textcolor{ForestGreen}{+ 0.75}  & 81.16 \textcolor{ForestGreen}{+ 0.36} \textcolor{red}{- 0.79} \textcolor{ForestGreen}{+ 0.44}  & 85.48 \textcolor{ForestGreen}{+ 0.34} \textcolor{ForestGreen}{+ 0.07} \textcolor{ForestGreen}{+ 0.54}  & 82.76 \textcolor{ForestGreen}{+ 0.47} \textcolor{red}{- 2.10} \textcolor{ForestGreen}{+ 0.75}  \\
& ter & 106.35 \textcolor{ForestGreen}{- 0.30} \textcolor{ForestGreen}{- 0.13} \textcolor{ForestGreen}{- 0.47}  & 99.34 \textcolor{ForestGreen}{- 1.04} \textcolor{ForestGreen}{- 0.21} \textcolor{ForestGreen}{- 0.94}  & 96.12 \textcolor{ForestGreen}{- 2.02} \textcolor{red}{+ 0.99} \textcolor{ForestGreen}{- 2.49}  & 101.99 \textcolor{ForestGreen}{- 0.03} \textcolor{ForestGreen}{- 0.07} \textcolor{ForestGreen}{- 0.39}  & 99.77 \textcolor{ForestGreen}{- 0.19} \textcolor{ForestGreen}{- 0.40} \textcolor{ForestGreen}{- 0.79}  \\

\cline{2-7}

\multirow{5}{*}{\texttt{aya-23-8B}} & BLEU & 14.28 \textcolor{ForestGreen}{+ 0.81} \textcolor{ForestGreen}{+ 0.28} \textcolor{ForestGreen}{+ 0.75}  & 14.50 \textcolor{ForestGreen}{+ 0.39} \textcolor{ForestGreen}{+ 0.19} \textcolor{ForestGreen}{+ 0.71}  & 24.49 \textcolor{ForestGreen}{+ 2.36} \textcolor{ForestGreen}{+ 0.08} \textcolor{ForestGreen}{+ 2.37}  & 9.22 \textcolor{ForestGreen}{+ 0.23} \textcolor{ForestGreen}{+ 0.35} \textcolor{ForestGreen}{+ 0.45}  & 16.39 \textcolor{ForestGreen}{+ 1.36} \textcolor{ForestGreen}{+ 0.61} \textcolor{ForestGreen}{+ 1.35}  \\
& ChrF & 14.77 \textcolor{ForestGreen}{+ 0.82} \textcolor{ForestGreen}{+ 0.28} \textcolor{ForestGreen}{+ 0.75}  & 15.12 \textcolor{ForestGreen}{+ 0.40} \textcolor{ForestGreen}{+ 0.20} \textcolor{ForestGreen}{+ 0.71}  & 25.11 \textcolor{ForestGreen}{+ 2.37} \textcolor{ForestGreen}{+ 0.07} \textcolor{ForestGreen}{+ 2.39}  & 9.68 \textcolor{ForestGreen}{+ 0.25} \textcolor{ForestGreen}{+ 0.37} \textcolor{ForestGreen}{+ 0.47}  & 16.78 \textcolor{ForestGreen}{+ 1.39} \textcolor{ForestGreen}{+ 0.62} \textcolor{ForestGreen}{+ 1.37}  \\
& ChrF++ & 14.77 \textcolor{ForestGreen}{+ 0.82} \textcolor{ForestGreen}{+ 0.28} \textcolor{ForestGreen}{+ 0.75}  & 15.12 \textcolor{ForestGreen}{+ 0.40} \textcolor{ForestGreen}{+ 0.20} \textcolor{ForestGreen}{+ 0.71}  & 25.11 \textcolor{ForestGreen}{+ 2.37} \textcolor{ForestGreen}{+ 0.07} \textcolor{ForestGreen}{+ 2.39}  & 9.68 \textcolor{ForestGreen}{+ 0.25} \textcolor{ForestGreen}{+ 0.37} \textcolor{ForestGreen}{+ 0.47}  & 16.78 \textcolor{ForestGreen}{+ 1.39} \textcolor{ForestGreen}{+ 0.62} \textcolor{ForestGreen}{+ 1.37}  \\
& COMET & 81.55 \textcolor{ForestGreen}{+ 1.03} \textcolor{red}{- 0.62} \textcolor{ForestGreen}{+ 1.23}  & 83.88 \textcolor{ForestGreen}{+ 0.68} \textcolor{ForestGreen}{+ 0.04} \textcolor{ForestGreen}{+ 0.80}  & 82.55 \textcolor{ForestGreen}{+ 1.22} \textcolor{red}{- 0.70} \textcolor{ForestGreen}{+ 1.23}  & 84.42 \textcolor{ForestGreen}{+ 0.81} \textcolor{ForestGreen}{+ 0.02} \textcolor{ForestGreen}{+ 1.26}  & 82.72 \textcolor{ForestGreen}{+ 1.14} \textcolor{red}{- 2.08} \textcolor{ForestGreen}{+ 1.29}  \\
& ter & 102.45 \textcolor{ForestGreen}{- 0.92} \textcolor{ForestGreen}{- 0.37} \textcolor{ForestGreen}{- 0.91}  & 95.18 \textcolor{ForestGreen}{- 0.69} \textcolor{ForestGreen}{- 0.42} \textcolor{ForestGreen}{- 0.97}  & 89.07 \textcolor{ForestGreen}{- 2.93} \textcolor{ForestGreen}{- 0.28} \textcolor{ForestGreen}{- 2.99}  & 100.30 \textcolor{ForestGreen}{- 0.43} \textcolor{ForestGreen}{- 0.58} \textcolor{ForestGreen}{- 0.72}  & 96.28 \textcolor{ForestGreen}{- 1.73} \textcolor{ForestGreen}{- 0.84} \textcolor{ForestGreen}{- 1.69}  \\

\cline{2-7}

\multirow{5}{*}{\texttt{gpt-4o-mini}} & BLEU & 14.37 \textcolor{ForestGreen}{+ 0.53} \textcolor{red}{- 0.42} \textcolor{ForestGreen}{+ 0.61}  & 17.21 \textcolor{ForestGreen}{+ 1.22} \textcolor{ForestGreen}{+ 1.39} \textcolor{ForestGreen}{+ 1.34}  & 24.45 \textcolor{ForestGreen}{+ 4.38} \textcolor{ForestGreen}{+ 1.28} \textcolor{ForestGreen}{+ 3.40}  & 10.55 \textcolor{ForestGreen}{+ 0.05} \textcolor{ForestGreen}{+ 0.03} \textcolor{ForestGreen}{+ 0.01}  & 18.02 \textcolor{ForestGreen}{+ 1.52} \textcolor{ForestGreen}{+ 0.40} \textcolor{ForestGreen}{+ 1.60}  \\
& ChrF & 14.82 \textcolor{ForestGreen}{+ 0.55} \textcolor{red}{- 0.42} \textcolor{ForestGreen}{+ 0.63}  & 17.93 \textcolor{ForestGreen}{+ 1.23} \textcolor{ForestGreen}{+ 1.39} \textcolor{ForestGreen}{+ 1.37}  & 25.04 \textcolor{ForestGreen}{+ 4.40} \textcolor{ForestGreen}{+ 1.31} \textcolor{ForestGreen}{+ 3.45}  & 10.99 \textcolor{ForestGreen}{+ 0.06} \textcolor{ForestGreen}{+ 0.03} \textcolor{ForestGreen}{+ 0.03}  & 18.44 \textcolor{ForestGreen}{+ 1.52} \textcolor{ForestGreen}{+ 0.40} \textcolor{ForestGreen}{+ 1.62}  \\
& ChrF++ & 14.82 \textcolor{ForestGreen}{+ 0.55} \textcolor{red}{- 0.42} \textcolor{ForestGreen}{+ 0.63}  & 17.93 \textcolor{ForestGreen}{+ 1.23} \textcolor{ForestGreen}{+ 1.39} \textcolor{ForestGreen}{+ 1.37}  & 25.04 \textcolor{ForestGreen}{+ 4.40} \textcolor{ForestGreen}{+ 1.31} \textcolor{ForestGreen}{+ 3.45}  & 10.99 \textcolor{ForestGreen}{+ 0.06} \textcolor{ForestGreen}{+ 0.03} \textcolor{ForestGreen}{+ 0.03}  & 18.44 \textcolor{ForestGreen}{+ 1.52} \textcolor{ForestGreen}{+ 0.40} \textcolor{ForestGreen}{+ 1.62}  \\
& COMET & 83.56 \textcolor{ForestGreen}{+ 0.86} \textcolor{red}{- 0.19} \textcolor{ForestGreen}{+ 0.93}  & 86.08 \textcolor{ForestGreen}{+ 0.25} \textcolor{red}{- 0.13} \textcolor{ForestGreen}{+ 0.07}  & 84.75 \textcolor{ForestGreen}{+ 0.33} \textcolor{red}{- 1.03} \textcolor{ForestGreen}{+ 0.13}  & 87.91 \textcolor{ForestGreen}{+ 0.16} \textcolor{red}{- 0.01} \textcolor{ForestGreen}{+ 0.19}  & 84.92 \textcolor{ForestGreen}{+ 0.44} \textcolor{red}{- 2.02} \textcolor{ForestGreen}{+ 0.49}  \\
& ter & 102.33 \textcolor{ForestGreen}{- 1.04} \textcolor{red}{+ 0.25} \textcolor{ForestGreen}{- 1.16}  & 93.01 \textcolor{ForestGreen}{- 2.23} \textcolor{ForestGreen}{- 2.45} \textcolor{ForestGreen}{- 2.24}  & 89.16 \textcolor{ForestGreen}{- 5.36} \textcolor{ForestGreen}{- 1.80} \textcolor{ForestGreen}{- 4.21}  & 98.39 \textcolor{ForestGreen}{- 0.07} \textcolor{ForestGreen}{- 0.13} \textcolor{ForestGreen}{- 0.06}  & 94.40 \textcolor{ForestGreen}{- 1.81} \textcolor{ForestGreen}{- 0.51} \textcolor{ForestGreen}{- 1.93}  \\

\cline{2-7}

\multirow{5}{*}{\texttt{nllb}} & BLEU & 15.42 \textcolor{red}{- 0.31} \textcolor{red}{- 0.77} \textcolor{ForestGreen}{+ 0.29}  & 10.24 \textcolor{ForestGreen}{+ 2.19} \textcolor{ForestGreen}{+ 2.07} \textcolor{ForestGreen}{+ 3.10}  & 22.68 \textcolor{ForestGreen}{+ 2.73} \textcolor{ForestGreen}{+ 0.90} \textcolor{ForestGreen}{+ 3.40}  & 8.24 \textcolor{ForestGreen}{+ 1.08} \textcolor{ForestGreen}{+ 0.88} \textcolor{ForestGreen}{+ 1.21}  & 19.18 \textcolor{red}{- 0.10} \textcolor{red}{- 0.24} \textcolor{ForestGreen}{+ 1.44}  \\
& ChrF & 15.95 \textcolor{red}{- 0.32} \textcolor{red}{- 0.77} \textcolor{ForestGreen}{+ 0.30}  & 10.82 \textcolor{ForestGreen}{+ 2.26} \textcolor{ForestGreen}{+ 2.16} \textcolor{ForestGreen}{+ 3.21}  & 23.33 \textcolor{ForestGreen}{+ 2.77} \textcolor{ForestGreen}{+ 0.91} \textcolor{ForestGreen}{+ 3.43}  & 8.61 \textcolor{ForestGreen}{+ 1.11} \textcolor{ForestGreen}{+ 0.93} \textcolor{ForestGreen}{+ 1.27}  & 19.63 \textcolor{red}{- 0.07} \textcolor{red}{- 0.20} \textcolor{ForestGreen}{+ 1.50}  \\
& ChrF++ & 15.95 \textcolor{red}{- 0.32} \textcolor{red}{- 0.77} \textcolor{ForestGreen}{+ 0.30}  & 10.82 \textcolor{ForestGreen}{+ 2.26} \textcolor{ForestGreen}{+ 2.16} \textcolor{ForestGreen}{+ 3.21}  & 23.33 \textcolor{ForestGreen}{+ 2.77} \textcolor{ForestGreen}{+ 0.91} \textcolor{ForestGreen}{+ 3.43}  & 8.61 \textcolor{ForestGreen}{+ 1.11} \textcolor{ForestGreen}{+ 0.93} \textcolor{ForestGreen}{+ 1.27}  & 19.63 \textcolor{red}{- 0.07} \textcolor{red}{- 0.20} \textcolor{ForestGreen}{+ 1.50}  \\
& COMET & 81.23 \textcolor{ForestGreen}{+ 1.28} \textcolor{red}{- 0.50} \textcolor{ForestGreen}{+ 1.99}  & 80.19 \textcolor{ForestGreen}{+ 1.61} \textcolor{ForestGreen}{+ 0.38} \textcolor{ForestGreen}{+ 2.84}  & 78.70 \textcolor{ForestGreen}{+ 3.81} \textcolor{red}{- 1.59} \textcolor{ForestGreen}{+ 1.62}  & 83.05 \textcolor{ForestGreen}{+ 1.71} \textcolor{ForestGreen}{+ 0.70} \textcolor{ForestGreen}{+ 2.55}  & 80.46 \textcolor{ForestGreen}{+ 2.80} \textcolor{red}{- 2.38} \textcolor{ForestGreen}{+ 1.63}  \\
& ter & 101.29 \textcolor{red}{+ 0.17} \textcolor{red}{+ 0.71} \textcolor{ForestGreen}{- 0.47}  & 101.05 \textcolor{ForestGreen}{- 2.45} \textcolor{ForestGreen}{- 2.59} \textcolor{ForestGreen}{- 3.46}  & 91.40 \textcolor{ForestGreen}{- 3.07} \textcolor{ForestGreen}{- 0.89} \textcolor{ForestGreen}{- 4.01}  & 101.41 \textcolor{ForestGreen}{- 1.31} \textcolor{ForestGreen}{- 0.86} \textcolor{ForestGreen}{- 1.28}  & 93.26 \textcolor{red}{+ 0.27} \textcolor{red}{+ 0.37} \textcolor{ForestGreen}{- 1.58}  \\

\cline{2-7}

\multirow{5}{*}{\texttt{seamless}} & BLEU & 15.38 \textcolor{ForestGreen}{+ 1.09} \textcolor{ForestGreen}{+ 0.45} \textcolor{ForestGreen}{+ 1.66}  & 13.67 \textcolor{ForestGreen}{+ 1.10} \textcolor{ForestGreen}{+ 0.73} \textcolor{ForestGreen}{+ 1.99}  & 24.34 \textcolor{ForestGreen}{+ 5.21} \textcolor{ForestGreen}{+ 1.49} \textcolor{ForestGreen}{+ 5.35}  & 9.42 \textcolor{ForestGreen}{+ 0.56} \textcolor{ForestGreen}{+ 0.42} \textcolor{ForestGreen}{+ 0.91}  & 18.43 \textcolor{ForestGreen}{+ 0.95} \textcolor{ForestGreen}{+ 0.35} \textcolor{ForestGreen}{+ 1.70}  \\
& ChrF & 15.90 \textcolor{ForestGreen}{+ 1.10} \textcolor{ForestGreen}{+ 0.46} \textcolor{ForestGreen}{+ 1.68}  & 14.34 \textcolor{ForestGreen}{+ 1.08} \textcolor{ForestGreen}{+ 0.72} \textcolor{ForestGreen}{+ 2.05}  & 25.05 \textcolor{ForestGreen}{+ 5.25} \textcolor{ForestGreen}{+ 1.50} \textcolor{ForestGreen}{+ 5.37}  & 9.91 \textcolor{ForestGreen}{+ 0.55} \textcolor{ForestGreen}{+ 0.41} \textcolor{ForestGreen}{+ 0.91}  & 18.88 \textcolor{ForestGreen}{+ 0.97} \textcolor{ForestGreen}{+ 0.37} \textcolor{ForestGreen}{+ 1.73}  \\
& ChrF++ & 15.90 \textcolor{ForestGreen}{+ 1.10} \textcolor{ForestGreen}{+ 0.46} \textcolor{ForestGreen}{+ 1.68}  & 14.34 \textcolor{ForestGreen}{+ 1.08} \textcolor{ForestGreen}{+ 0.72} \textcolor{ForestGreen}{+ 2.05}  & 25.05 \textcolor{ForestGreen}{+ 5.25} \textcolor{ForestGreen}{+ 1.50} \textcolor{ForestGreen}{+ 5.37}  & 9.91 \textcolor{ForestGreen}{+ 0.55} \textcolor{ForestGreen}{+ 0.41} \textcolor{ForestGreen}{+ 0.91}  & 18.88 \textcolor{ForestGreen}{+ 0.97} \textcolor{ForestGreen}{+ 0.37} \textcolor{ForestGreen}{+ 1.73}  \\
& COMET & 81.96 \textcolor{ForestGreen}{+ 1.18} \textcolor{red}{- 0.39} \textcolor{ForestGreen}{+ 1.72}  & 80.70 \textcolor{ForestGreen}{+ 2.18} \textcolor{ForestGreen}{+ 0.16} \textcolor{ForestGreen}{+ 2.95}  & 83.76 \textcolor{ForestGreen}{+ 0.97} \textcolor{red}{- 0.94} \textcolor{ForestGreen}{+ 0.88}  & 83.70 \textcolor{ForestGreen}{+ 0.88} \textcolor{ForestGreen}{+ 0.10} \textcolor{ForestGreen}{+ 1.81}  & 83.12 \textcolor{ForestGreen}{+ 1.50} \textcolor{red}{- 1.79} \textcolor{ForestGreen}{+ 1.70}  \\
& ter & 101.34 \textcolor{ForestGreen}{- 1.48} \textcolor{ForestGreen}{- 0.77} \textcolor{ForestGreen}{- 2.19}  & 97.26 \textcolor{ForestGreen}{- 1.76} \textcolor{ForestGreen}{- 1.07} \textcolor{ForestGreen}{- 2.62}  & 89.75 \textcolor{ForestGreen}{- 6.47} \textcolor{ForestGreen}{- 2.12} \textcolor{ForestGreen}{- 6.61}  & 100.59 \textcolor{ForestGreen}{- 1.35} \textcolor{ForestGreen}{- 1.15} \textcolor{ForestGreen}{- 1.77}  & 94.15 \textcolor{ForestGreen}{- 1.03} \textcolor{ForestGreen}{- 0.38} \textcolor{ForestGreen}{- 2.03}  \\

\bottomrule
\end{tabular}

}
\caption{Full evaluation results across five models and five languages using BLEU, ChrF, ChrF++, COMET, and TER. The first black value in each column represents direct translation scores. The second, third, and fourth values (colored \textcolor{red}{red} or \textcolor{ForestGreen}{green}) indicate the relative change in performance when applying the prompting-powered refinement method, the word alignment method, and post-hoc prompting after word alignment for improved morphological and grammatical correctness, respectively, compared to direct translation. Lower TER scores indicate better alignment with the reference.}

\label{tab:exp_results_full}
\end{table*}

\clearpage

\clearpage

\end{document}